\documentclass[review]{elsarticle}
\usepackage{amssymb}
\usepackage{amsmath}
\usepackage{lineno}
\usepackage{hyperref}

\newdefinition{definition}{Definition}

\begin{document}

\begin{frontmatter}

\title{Multilayer deep feature extraction for visual texture recognition}

\author[ime]{Lucas O. Lyra\corref{cor1}}
\cortext[cor1]{Corresponding author}
\ead{lucas.oliveira.lyra@alumni.usp.br}
\author[ime]{Antonio Elias Fabris}
\ead{fabris@usp.br}
\author[imecc]{Joao B. Florindo}
\ead{jbflorindo@ime.unicamp.br}

\address[ime]{Institute of Mathematics and Statistics - University of S\~{a}o Paulo\\
Rua do Mat\~{a}o, 1010 - Butant\~{a}, CEP 05508-090, São Paulo, SP, Brasil}

\address[imecc]{Institute of Mathematics, Statistics and Scientific Computing - University of Campinas\\
Rua S\'{e}rgio Buarque de Holanda, 651 - Bar\~{a}o Geraldo, CEP 13083-859, Campinas, SP, Brasil}

\begin{abstract}
Convolutional neural networks have shown successful results in image classification achieving real-time results superior to the human level. However, texture images still pose some challenge to these models due, for example, to the limited availability of data for training in several problems where these images appear, high inter-class similarity, the absence of a global viewpoint of the object represented, and others.
In this context, the present paper is focused on improving the accuracy of convolutional neural networks in texture classification. This is done by extracting features from multiple convolutional layers of a pretrained neural network and aggregating such features using Fisher vector.
The reason for using features from earlier convolutional layers is obtaining information that is less domain specific. 
We verify the effectiveness of our method on texture classification of benchmark datasets, as well as on a practical task of Brazilian plant species identification.
In both scenarios, Fisher vectors calculated on multiple layers outperform state-of-art methods, confirming that early convolutional layers provide important information about the texture image for classification.
\end{abstract}



\end{frontmatter}

\section{Introduction}

Texture is one of the most important image attributes in computational vision. It provides information on the spatial arrangement of the pixel intensities in an image. Textures can be useful for recognition of material properties, specially when other image attributes, like shape, are not useful. They play an important role in remote sensing \cite{ansari2020urban}, material science \cite{nurzynska2020application}, medicine \cite{scalco2017texture}, agriculture \cite{jana2017automatic} and many other fields.

The problem of recognizing a texture can be divided into two tasks: the first is extracting features from an image; the second one is training a classifier for feature recognition.
Given the way those tasks are performed, techniques can be divided into two groups. In the first case, features are extracted by a computer vision method and used by a classical machine learning algorithm for classification. In the other one, both feature extraction and classification are performed by a deep neural network, usually Convolutional Neural Network (CNN). In that case, parameters are learned by an optimization algorithm.

Although CNNs have been quite successful for image classification, textures are still challenging. This is a consequence of the limited availability of data for training in areas of application, such as medicine, for example, and other characteristics such as the high inter-class similarity and the lack of a global viewpoint over the analyzed object. 
Even if we consider transferring knowledge from large databases, like ImageNet, there can be significant domain shift between those large databases and the field of research interest. In this context, the literature has presented a growing number of studies combining CNNs with classical texture descriptors \cite{gibert2018classification,wan2019information,liu2019multi}.

When using classical texture features, the performance classification of algorithms rely heavily on how well the extracted features describe the image. Features extracted from the last convolutional layer of CNNs tend to be better than features extracted by classical filter banks \cite{cimpoi2016deep}. However, given the domain shift mentioned in the previous paragraph, features from the last convolutional layer might be too specific to the training database. 

In this context, we propose a method that combines generalist local features with specific ones into a single set of features. In order to evaluate such method, we compute Fisher Vectors on this set of features and classify them using Support Vector Machine (SVM).
The major contributions of this paper are the following:
\begin{enumerate}
    \item Up to our knowledge, this is the first time Fisher Vectors are associated with features extracted from multiple layers of a CNN;
    \item We propose the application of normalization on descriptors extracted from fully-connected layers and evaluate the impact of the proposed normalization in classification accuracy;
    \item We obtain results competitive with other methods available in literature, establishing, up to our knowledge, new state-of-the-art performance in Flickr Material Database (FMD) \cite{sharan2009material}, Describable Textures Dataset (DTD) \cite{cimpoi2014describing} and in Brazilian plant species 1200Tex database \cite{CMB09}.
\end{enumerate}

In Section~\ref{sec:related}, we mention and briefly describe some related works. In Section~\ref{sec:background} the theoretical background necessary for the presentation of the proposed method is described, with Section~\ref{sec:deep} giving a brief general description of CNNs and Section~\ref{sec:fisher} focusing on how Fisher Kernels can be used in texture descriptors. In Section~\ref{sec:proposed} we present the proposed method for visual texture classification. Section~\ref{sec:experiments} shows our procedures to test and validate the performance of our method. In Section~\ref{sec:results} we present and discuss the obtained results. Finally, Section~\ref{sec:conclusions} presents the general conclusions of our research. 
The code will be available at \url{https://github.com/lolyra/multilayer}.

\section{Related works}
\label{sec:related}

Earlier works on texture recognition were based on using handcrafted features that are invariant to scale, illumination and translation. 
Scale Invariant Feature Transform (SIFT) \cite{lowe2004distinctive}, Local Binary Patterns (LBP) \cite{ojala2002multiresolution} and variants \cite{hafiane2015joint, ruichek2018local} are prominent examples in this regard in the literature. 

On top of those handcrafted feature extractors, an encoder is needed to combine features into a single descriptor vector that can be used in a discriminative classifier.
Traditional encoders include Bag-of-Visual Words and its variations \cite{malik2001contour,lazebnik2005sparse, zhang2007local, sharan2013recognizing}, Vector of Locally Aggregated Descriptors (VLAD) \cite{amato2013large} and Fisher Vectors (FV) \cite{perronnin2007fisher, perronnin2010improving}.

However, in recent works, a shift has been made from handcrafted feature extractors to deep neural networks. Since texture recognition databases are frequently very small to train deep neural networks from scratch, most of the proposed methods use pretrained CNNs on large databases, like ImageNet. This is the case of Cimpoi et al. \cite{cimpoi2016deep}. 
They proposed a method combining CNNs with traditional encoders that achieved state-of-the-art results. However, its good performance requires the use of multiple scales in the input image, which implies using CNNs several times. 

More recently, improvements on the association of CNNs with traditional encoders have been proposed. 
Song et al.~\cite{song2017locally} proposed a method that consists of optimizing the Fisher Vector for classification by applying a simple neural network on top of the FV descriptor and training it from scratch.
Lin et. al.~\cite{lin2017bilinear} avoid the use of generative models such as GMM, applying outer product to features extracted from two neural networks, thus obtaining second-order image features that are used to calculate FV or VLAD.

Given the overall good performance of SIFT even when compared to CNNs, a step towards a hybrid model was taken by Jbene et al. \cite{jbene2019fusion}. They calculate Fisher Vectors on features extracted by CNN and on features extracted by SIFT, later combining for classification purposes.

In addition, another source of features that can contribute to a good performance for classification are those extracted from other layers of a CNN rather than only the last convolutional layer. 
Such approach is presented by Chen et at. \cite{chen2021deep} where encoding is performed by calculating statistical self-similarity using a soft histogram of local differential box-counting dimensions of cross-layer features.

More recent works have focused on alternatives to traditional encoders, like FV, either to create an end-to-end trainable model or reduce computational costs. Florindo et al. \cite{florindo2021using} performs aggregation by joining two different fully-connected layers output . The first one calculated over original image and the other one calculated over an entropy measure of the original image. In other work \cite{florindo2021visgraphnet}, encoding is performed by visibility graphs.

In the scope of end-to-end training, Mao et al. \cite{mao2021deep} obtain a fully trainable model by performing encoding with an aggregation module, which consists of convolutions and average pooling. In Xu et al. \cite{xu2021encoding}, encoding is performed by a local-global hierarchical fractal analysis.

\section{Background}
\label{sec:background}

In this section, we describe the concepts needed to understand the proposed model. In Section~\ref{sec:deep}, we set the basic theory and describe the functioning of Convolutional Neural Networks (CNN), detailing some frequently used layers. In Section~\ref{sec:fisher}, we present a concise summary of Fisher Vector (FV).

\subsection{Deep Convolutional Features}
\label{sec:deep}

A CNN is a neural network usually developed to handle images. Nodes in each layer can be organized in a multi-dimensional space. Using three dimensions, for example, it is possible to explore relations among neighbor pixels and among color channels.

This type of neural network can be decomposed into two main parts. The first one is used for extracting features from images. It is usually composed by convolutional, pooling, activation and normalization layers. The second part is composed by fully-connected layers whose purpose is classification.

Classical extraction of features is performed by applying convolutional filters to the input image \cite{leung2001representing}. In this context, the feature extraction part of a CNN can be seen as a bank of filters, where each channel from each convolutional layer is a particular filter.

\subsubsection{Convolution Layer}
A convolutional layer is an appropriate mechanism to reduce the number of parameters and explore relationships between neighbor pixels. It is composed by multiple 2-dimensional kernels $K_{c}(i,j)=w_{i,j}$. Given a 2-dimensional input $I(i,j)$, each output $S_{c}(i,j)$ is the convolution of $I$ by $K_{c}$, which is given by:
\begin{equation}
    S_{c}(i,j)=\sum_{m=0}^{d-1}\sum_{n=0}^{d-1}I(i\cdot s-m,j\cdot s-n)K_{c}(m,n).
\end{equation}
The parameter $d$ is called \emph{kernel size} and $s$ is the \emph{stride}. Note that, generally, kernels have width equals height. Stride is the convolution step size.

Let $W$ and $H$ be, respectively, the width and height of $I$. As $i,j$ are not defined outside the set $[0,W-1]\times[0,H-1]$, $S$ has dimensions smaller than $I$. In order to increase output size, a parameter $p$, called $padding$, can be introduced. In such a case, we define $I(i,j)=0$ for $i\in [-p,0]\cup [W,W+p]$ and $j\in [-p,0]\cup [H,H+p]$. Thus, the output dimension $D_{out}$ is given by
\begin{equation}
D_{out} = \frac{D_{in}-d+2p}{2}-1,
\end{equation}
where $D_{in}$ can be either $W$ or $H$.
An activation function $f:\mathbb{R}\to\mathbb{R}$ is usually applied to $S$ in order to introduce non-linearity to the objective function estimator. One of the most popular activation function is called \emph{rectifier linear unit} and is given by
\begin{equation}
f(x) = \max(0,x).
\end{equation}

\subsubsection{Pooling Layer}
A pooling layer is used to reduce the number of parameters to be learned and the computational cost of the network. This is performed by reducing the input dimensions and helps preventing overfitting. In this layer, the input $I$ is reduced by merging a set of $m\times n$ pixels into a single one.
In general, pixels are combined by retrieving their maximum value. Thus the output $S$ is given by
\begin{equation}
S(i,j)=\max_{0\leq k \leq m}(\max_{0\leq l\leq n}(I(i\cdot m+k,j\cdot n+l))).
\end{equation}

\subsubsection{Dropout Layer}
\emph{Dropout} is a regularization technique used to avoid overfitting. It was introduced by Srivastava, et al. in \cite{srivastava2014dropout} and consists of avoiding the update of randomly selected neurons during one epoch. In the introductory paper, it is shown that the use of this technique has increased the accuracy of supervised learning tasks in areas such as computer vision, voice recognition and computational biology.

\subsubsection{Normalization Layer}
In neural network training, updates to the weights in early layers can change data distribution significantly in later layers. 
 This phenomenon is called \emph{internal covariate shift} and can make the training process very slow. 
In order to avoid it, a normalization layer is introduced. Using this layer, input data distribution can be imposed to have mean $0$ and variance $1$. 
Normalization can not only speed up the training process, but also act as a regularization layer, dismissing the need of a Dropout layer.

As noted in \cite{ioffe2015batch}, normalizing all data can be costly and a better approach would be batch normalization of the data. Thus let $m$ be the batch size and $d$ the dimension of the input. Let $x^{i}_{j}$ denote the $j$-th coordinate of the $i$-th input data from a batch. The normalization is given by
\begin{equation}
\hat{x}^{i}_{j} = \frac{x^{i}_{j}-\mu_{j}}{\sqrt{\sigma_{j}+\epsilon}},
\end{equation}
where $\mu_j$ and $\sigma_j$ denote, respectively, the mean and variance of the $j$-th components of the batch and $\epsilon$ is a positive constant to assure numerical stability.

\subsubsection{Fully-Connected Layer}
A fully-connected (FC) layer explores relations among all the components of the input data. In such layer, the multi-dimensional data from the previous one is rearranged into a one-dimensional vector $V$. The layer's output $S$ is also a one-dimensional vector and is given by
\begin{equation}
s_{j}=\sum_{i=0}^{d-1}w_{i,j}v_i,
\end{equation}
where $s_j$ is the $j$-th component of $S$ and $v_i$ is the $i$-th component of $V$, $d$ is the number of components of $V$ and $w$'s are the weights of the layer.

The FC layer is generally placed on top of the network to accomplish the classification task. Softmax function is normally used as an activation function after the last layer. The objective function guiding the optimization of the network is called \emph{loss function}. Some commonly used loss functions in classification task are \emph{cross-entropy} and \emph{Hinge} loss.

\subsection{Fisher Vector}
\label{sec:fisher}

Let $X = \{x_d,d=1\cdots D\}$ denote a sample of $D$ observations, $x_d\in\mathbb{R}^{N}$. Assume that the generation process of $X$ can be modeled by the probability density function $u_\lambda$ with parameters $\lambda$. Then one can characterize the observations in $X$ by the following gradient vector

\begin{equation}
\label{eq:grad}
G_{\lambda}^{X} = \nabla_{\lambda} \log u_{\lambda}(X).
\end{equation}

\noindent
The gradient vector given by Equation~(\ref{eq:grad}) can be classified using any classification algorithm. In \cite{jaakkola1998exploiting}, the Fisher information matrix $F_{\lambda}$ is suggested for this purpose:

\begin{equation}
F_{\lambda} = E_{X}[G_{\lambda}^{X} {G_{\lambda}^{X}}'].
\end{equation}
From this observation, a Fisher Kernel (FK) to measure similarity between two samples $X$ and $Y$ was proposed. Such kernel is defined by:

\begin{equation}
K_{FK}(X,Y) = {G_{\lambda}^{X}}'F_{\lambda}^{-1}G_{\lambda}^{Y}.
\end{equation}

As $F_{\lambda}^{-1}$ is positive semi-definite, so is $F_{\lambda}$. Using the Cholesky decomposition $F_{\lambda}^{-1} = {L_{\lambda}}'L_{\lambda}$, the FK can be re-written as:
\begin{equation}
K_{FK}(X,Y) = {\mathcal{G}_{\lambda}^{X}}'\mathcal{G}_{\lambda}^{Y}
\end{equation}
where
\begin{equation}
\label{eq:fisher}
\mathcal{G}_{\lambda}^{X} = L_{\lambda}\nabla_{\lambda} \log u_{\lambda}(X).
\end{equation}

The vector $\mathcal{G}_{\lambda}^{X}$ is called \emph{Fisher Vector} (FV). We have that FV $\mathcal{G}_{\lambda}^{X}$ and $G_{\lambda}^{X}$ have the same dimensionality \cite{sanchez2013image}. Therefore, we can conclude that performing classification with a linear kernel machine using an FV as feature vector is equivalent to performing a non-linear kernel machine using $K_{FK}$ as kernel.

\section{Proposed method}
\label{sec:proposed}

Here we propose an approach to use information from multiple layers of a CNN and Fisher vector encoding to perform classification. The current section is divided into two subsections. In Subsection \ref{sec:extraction}, we show the proposed strategy to build feature vectors. In Subsection \ref{sec:classification}, we show the classification process using such vectors.

\begin{figure}
    \centering
    \includegraphics[width=1\textwidth]{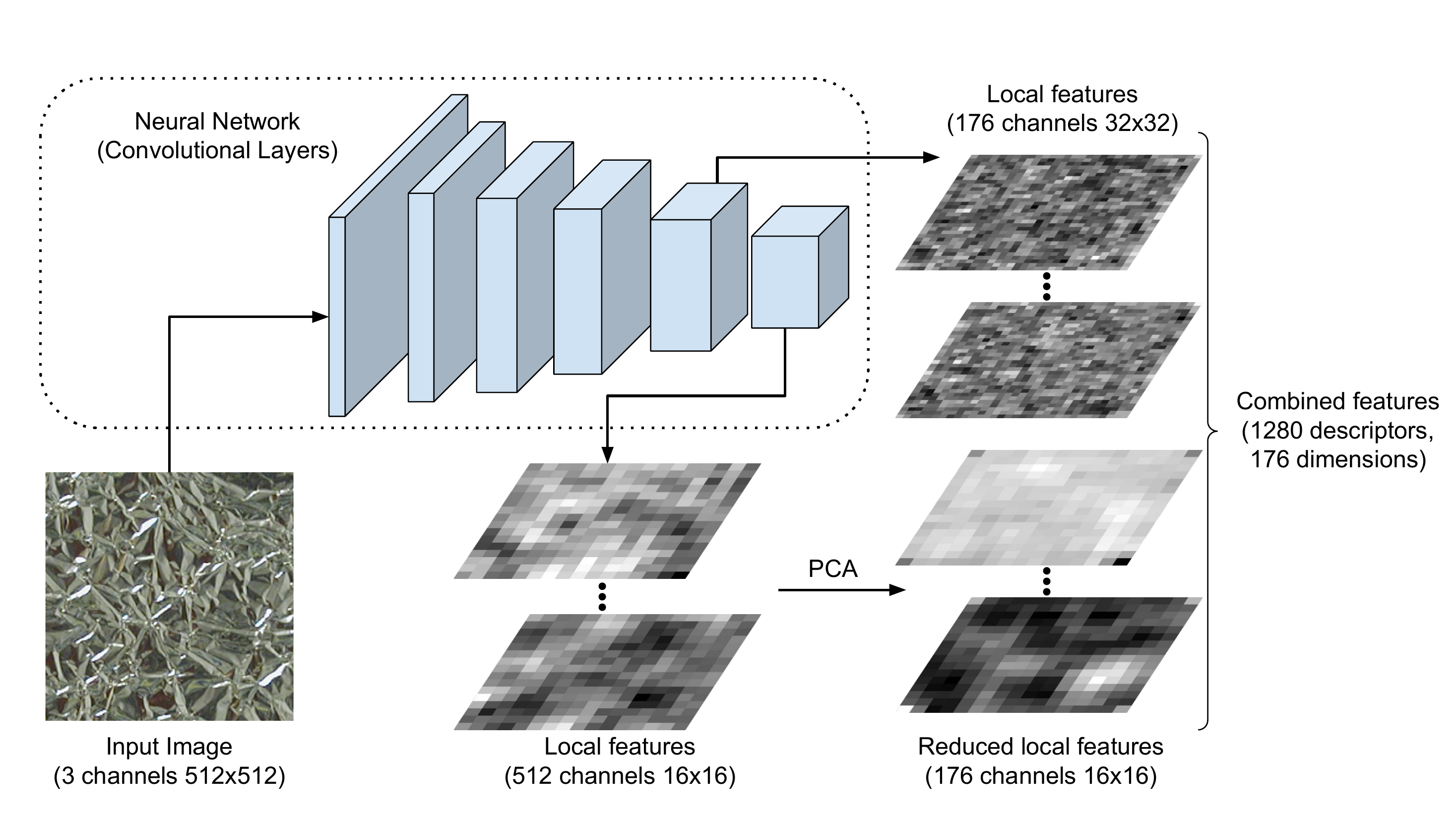}
    \caption{Feature extraction with the proposed method. From left to right we have the input texture, convolutional layers and local features extracted from the last two layers.}
    \label{fig:schema}
\end{figure}

\subsection{Feature Extraction}
\label{sec:extraction}
In the first stage of our methodology, we are interested in using Fisher vectors to describe information extracted from multiple layers of a convolutional neural network. Initially, we take a CNN architecture pretrained on ImageNet and use it as a feature extractor. We present the texture image as input to the pretrained CNN and collect the outputs of the last and penultimate convolutional layers. Both layers contain feature information about the image, the last layer presenting more high-level information than the previous one. 

\begin{definition}
Let the set $X_n = \{x_{t},t=1\cdots T_n \mid x_t\in\mathbb{R}^{D_n} \}$ denote the output of the $n$-th convolutional layer, where $T_n = W_n \times H_n$ is the resolution of each channel and $D_n$ is the number of channels. We call $x \in X_n$ a local feature and $X_n$ a set of local features extracted from the $n$-th convolutional layer.
\end{definition}

Let $N$ denote the number of convolutional layers in a CNN. Our method takes the local feature sets $X_{N-1}$ and $X_{N}$. We are interested in creating a single set $X$ of local features, but we usually have $T_{N-1} > T_{N}$ and $D_{N-1} \leq D_{N}$. Thus, in order to combine $X_{N-1}$ and $X_{N}$, we apply Principal Component Analysis (PCA) \cite{abdi2010principal} to each element $x \in X_{N}$, so that the element with reduced dimension $x^{\prime}$ is such that $x^{\prime}\in\mathbb{R}^{D_{N-1}}$. We end up with a set $X=\{ x_t, t=1\cdots T\mid x_t \in \mathbb{R}^{D} \}$, where $D=D_1$ and $T=T_1+T_2$.
The proposed schema for feature extraction is exemplified in Figure~\ref{fig:schema}, where the neural network architecture used is EfficientNet-B5 \cite{tan2019efficientnet}.

Once we have a set of local features, we calculate the Fisher vector. In order to do so, we assume that the local features $x_l$ are generated independently by the distribution $u_{\lambda}$. Thus Equation~(\ref{eq:fisher}) becomes:

\begin{equation}
\mathcal{G}_{\lambda}^{X} = L_{\lambda}\frac{1}{T}\sum_{t=1}^{T}\nabla_{\lambda} \log u_{\lambda}(x_t).
\end{equation}

We choose $u_{\lambda}$ to be a Gaussian Mixture Model (GMM) composed by $K$ Gaussian distributions, that is,
\begin{equation}
u_{\lambda}(x) = \sum_{i=1}^{K}w_i u_i(x),
\end{equation}
where $\lambda=\{w_i,\mu_i,\Sigma_i,i=1,\cdots,K\}$ and $w_i$, $\mu_i$, $\Sigma_i$ denote, respectively, the weight, mean and covariance matrix associated with Gaussian $u_i$. 

Let $\gamma_t(i)$ denote the probability of an observation $x_t$ to be generated by the Gaussian $u_i$:
\begin{equation}
\gamma_i(x_t) = \frac{w_i u_i(x_t)}{\sum_{j=1}^{K} w_j u_j(x_t)}.
\end{equation}
We assume that covariance matrices are diagonal given that any distribution can be approximated with an arbitrary precision by a weighted sum of Gaussians with diagonal covariances \cite{perronnin2007fisher}. We denote $\sigma_i^{2} = \mathrm{diag}(\Sigma_i)$. Using the values of $L_\lambda$ and $\nabla_{\lambda} \log u_{\lambda}(X)$ derived in \cite{perronnin2007fisher}, we can rewrite Equation~(\ref{eq:fisher}) as:

\begin{align}
    \mathcal{G}_{w_i^d}^{X} &= \frac{1}{T\sqrt{w_i}}\sum_{t=1}^{T} 
    \left( \gamma_i(x_t)- w_i\right), \\
    \mathcal{G}_{\mu_i^d}^{X} &= \frac{1}{T\sqrt{w_i}}\sum_{t=1}^{T} \gamma_i(x_t)\left(\frac{x_t^d-\mu_i^d}{\sigma_i^d}\right), \\
    \mathcal{G}_{\sigma_i^d}^{X} &= \frac{1}{T\sqrt{2w_i}}\sum_{t=1}^{T} \gamma_i(x_t)\left[\frac{(x_t^d-\mu_i^d)^2}{(\sigma_i^d)^2}-1\right].
\end{align}

\subsection{Classification}
\label{sec:classification}

In our second stage, we extract information from fully-connected layers by removing the classification layer of the CNN. We are left with a feature vector that we call FC.

Before doing classification, we perform a transformation over both FV and FC features. Let $\mathbf{x}$ be a feature vector and let $x$ denote an element of $\mathbf{x}$. We apply power and $L_2$ normalization to $\mathbf{x}$, which can be written as:

\begin{align}
x \leftarrow &\mathrm{sign}(x)\sqrt{|x|}, \label{eq:norm1} \\
x \leftarrow &\frac{x}{\lVert \mathbf{x} \rVert} \label{eq:norm2}.
\end{align}
These transformations were proposed in \cite{perronnin2010improving} as a way to improve classification with Fisher vectors. We noticed that those transformations are also beneficial for classification in the case of FC. Once we have normalized feature vectors, we perform classification with Support Vector Machine (SVM), using a modified version of Bhattacharyya coefficient given in Definition~\ref{def:modified} as kernel.

\begin{definition}
\label{def:modified}
Let $\mathbf{x},\mathbf{y}\in\mathbb{R}^{N}$, the modified Bhattacharyya coefficient is given by the following measure of distance:
\begin{equation}
K(\mathbf{x},\mathbf{y}) = \sum_{i=1}^{N}\mathrm{sign}(x_i y_i)\sqrt{|x_i y_i|}.
\end{equation}
\end{definition}
Note that the modified Bhattacharyya coefficient can be rewritten as 
\begin{equation}
K(\mathbf{x},\mathbf{y}) = \phi(\mathbf{x})^{T}\phi(\mathbf{y}),
\end{equation}
where $\phi(\mathbf{x})$ is a vector whose coordinates are given by
\begin{equation}
\label{eq:phi}
\phi(\mathbf{x})_i = \mathrm{sign}(x_i)\sqrt{|x_i|}.
\end{equation}
Thus, we apply the transformation given by Equation~(\ref{eq:phi}) to the normalized feature vectors and proceed to classification with a linear SVM.

\begin{figure}
    \centering
    \includegraphics[width=1\textwidth]{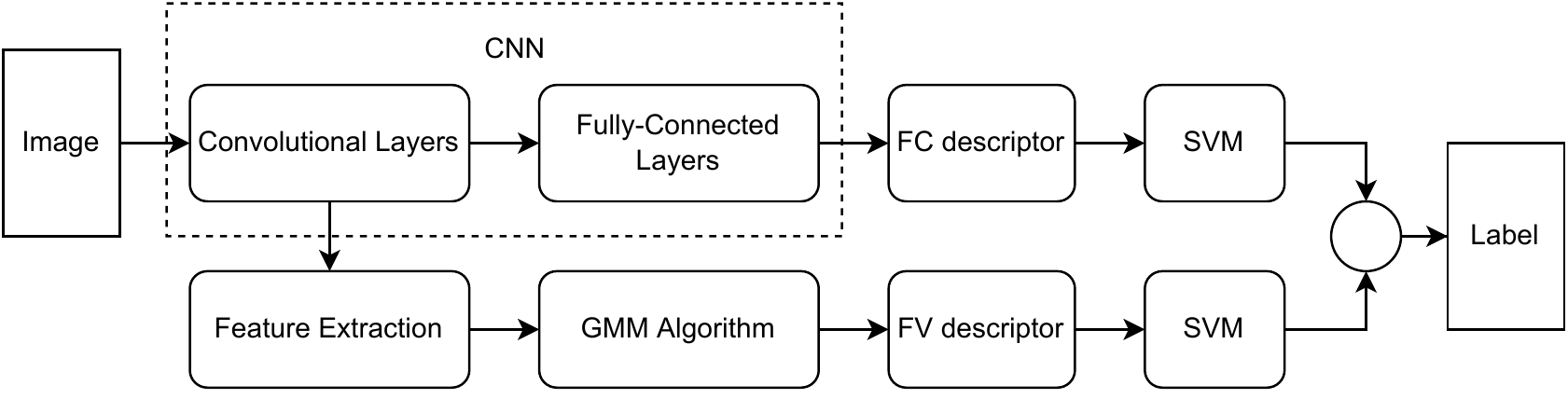}
    \caption{Summary of our proposed method for classifying a given image. Normalization is applied to FC and FV descriptors before classification with SVM.}
    \label{fig:summary}
\end{figure}

Finally, we combine classification with SVM trained on FC and FV data by applying soft assignment. Let $f_{\mathrm{FC}}$ denote the decision function of SVM trained on FC and $f_{\mathrm{FV}}$ the decision function of SVM trained on FV. Given FC $\mathbf{x}$ and FV $\mathbf{y}$ calculated on the same sample, we assign a class $c$ to the sample by:
\begin{equation}
c = \mathrm{argmax}(f_{\mathrm{FC}}(\mathbf{x}) + f_{\mathrm{FV}}(\mathbf{y})).
\end{equation}
\noindent
The proposed method for classification is summarized in Figure~\ref{fig:summary}.

\section{Experiments}
\label{sec:experiments}

In this section we describe how we evaluate our proposed methodology. We start by evaluating the effects of hyperparameters on classification accuracy. Our base model uses the EfficientNet-B5 architecture \cite{tan2019efficientnet} with pre-trained ImageNet weights, input image resolution of $512 \times 512$ and $64$ Gaussian distributions to model $u_\lambda$. Any hyperparameter change keeps the remaining parameters constant.

Fisher Vector accuracy can be affected by the number of Gaussian distributions that we use to model $u_\lambda$. Thus, using our base model, we tested the effect of this variation by reducing the number of Gaussian distributions. Another hyperparameter of our model is the input image resolution. We tested its effect on accuracy by downsampling the image in our base model. Also, we change the CNN architecture to see how our method behaves on other architectures.

Afterwards, we evaluated the effects of normalization of FC by comparing it with a model without normalization. Finally, we compare our base model with alternative state-of-art approaches. We conclude our experiments by applying our model to a practical task that consists in the identification of Brazilian plant species based on the scanned image of the leaf surface.

The databases used for method evalution are KTH-TIPS2-b, FMD, DTD, UIUC, UMD. The database used in our practical task is 1200Tex. All these databases are described in the following paragraphs.

KTH-TIPS2-b \cite{caputo2005class}, here referred to as KTH-TIPS, consists of 4 samples of images from 11 materials. Each sample is presented in 9 different scales, 3 poses and 4 lighting conditions. This represents a total of 108 images of 200x200 size per material per sample. In each round, we use 1 sample for training and 3 samples for testing.

FMD \cite{sharan2009material} consists of 10 classes containing 100 images each. Each image has a size of 512x384. We run 10 training/testing rounds, each randomly selecting half of the database for training and using the other half for testing.

DTD \cite{cimpoi2014describing} consists of 5640 images with varying sizes divided into 47 categories. This results in 120 images per class, which are divided into three equal parts: training, validation and testing. The database contains 10 splits of the data. For each one, we use training and validation parts for adjusting our model and the remaining part for testing.

UMD \cite{xu2009viewpoint} consists of 25 classes containing 40 images each. All images have a dimension of 1280x960. We evaluated our model 10 times in this dataset, each time randomly choosing 20 images from each class for training and the remainder for testing, following the same protocol as FMD.

UIUC \cite{lazebnik2005sparse}, as UMD, consists of 1000 images evenly divided in 25 classes. Each image has resolution of 640x480. In order to evaluate our method in this dataset, we use the same protocol applied to FMD.

1200Tex \cite{CMB09} consists of 1200 leaf surface images of 20 Brazilian plant species (classes). Each class contains 60 samples. We applied the same protocol followed in FMD to choose training and testing datasets.

\section{Results and Discussion}
\label{sec:results}

\begin{figure}[!htpb]
	\centering
	\begin{tabular}{cc}
		KTH-TIPS & FMD\\
		\includegraphics[width=.45\textwidth]{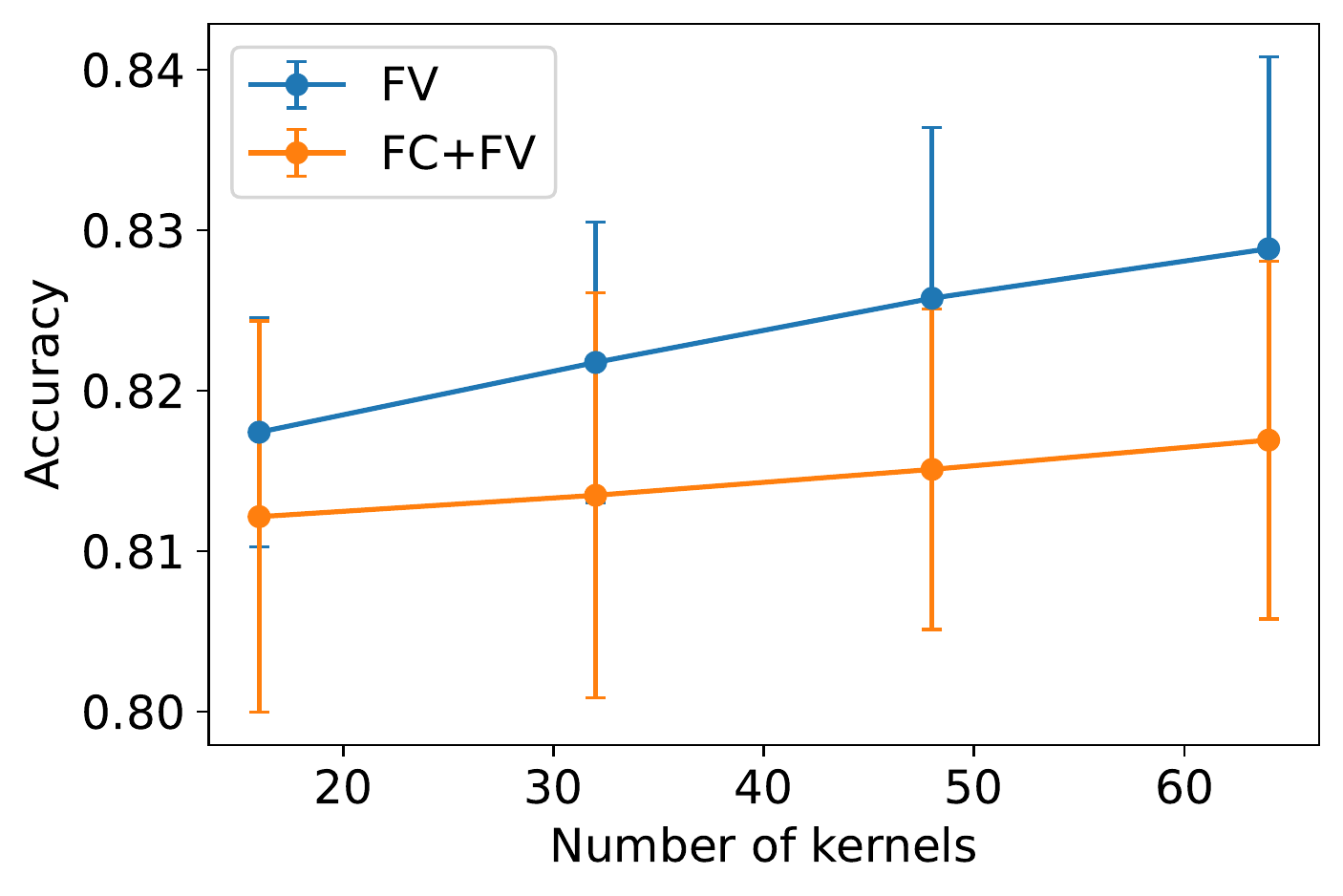} &
		\includegraphics[width=.45\textwidth]{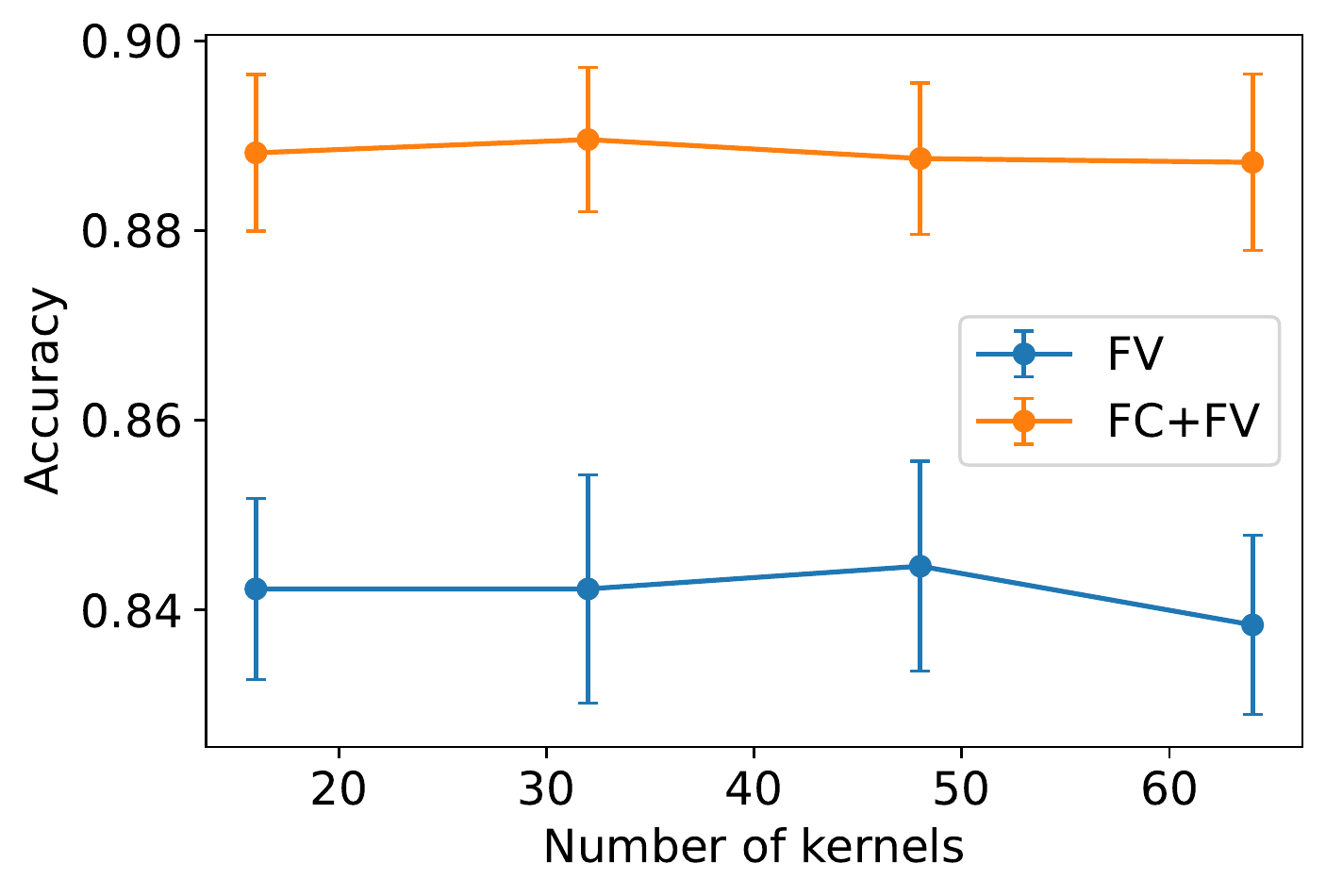}\\
	\end{tabular}
	\begin{tabular}{cc}
		DTD & UMD\\
		\includegraphics[width=.45\textwidth]{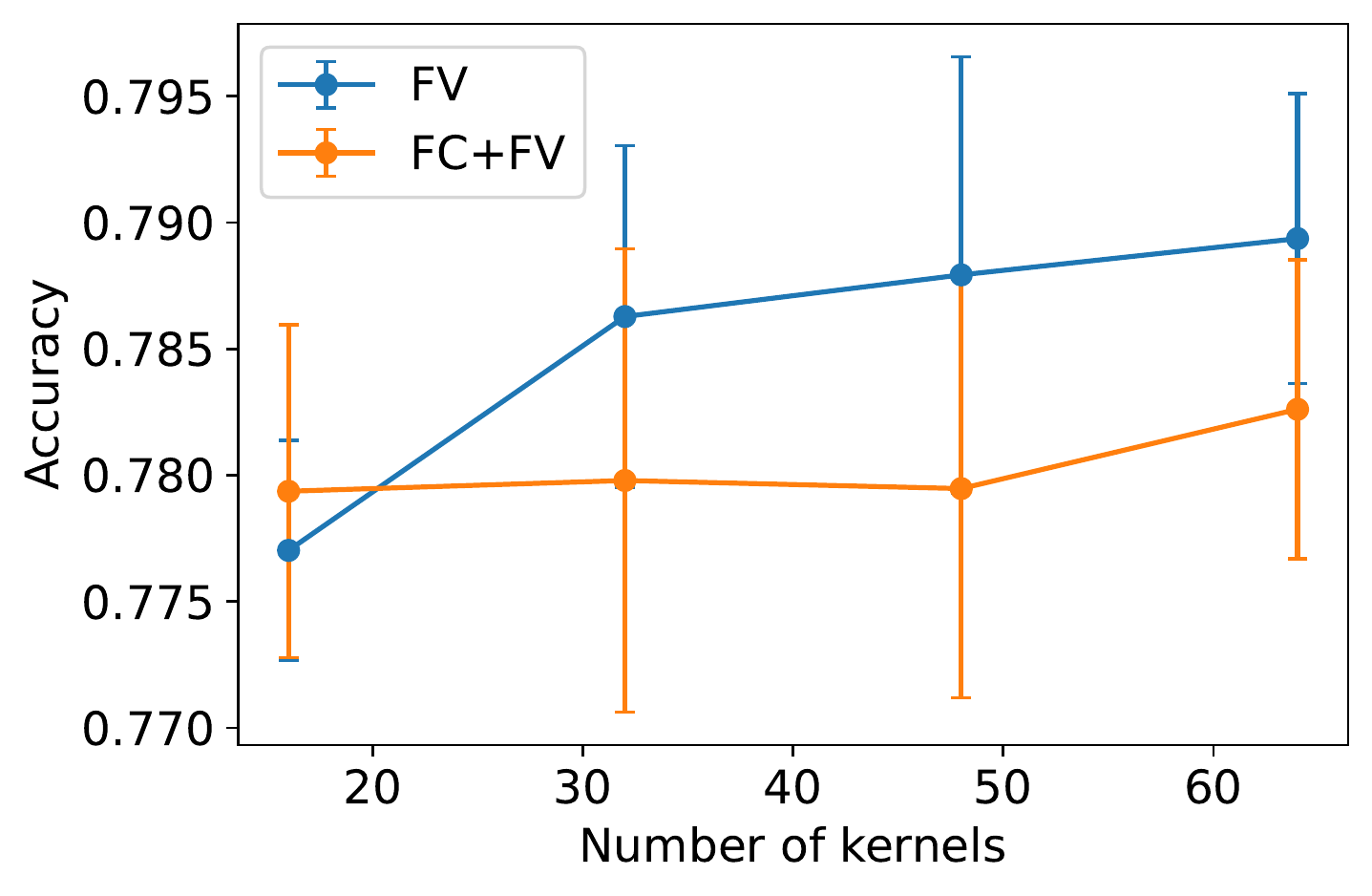} &
		\includegraphics[width=.45\textwidth]{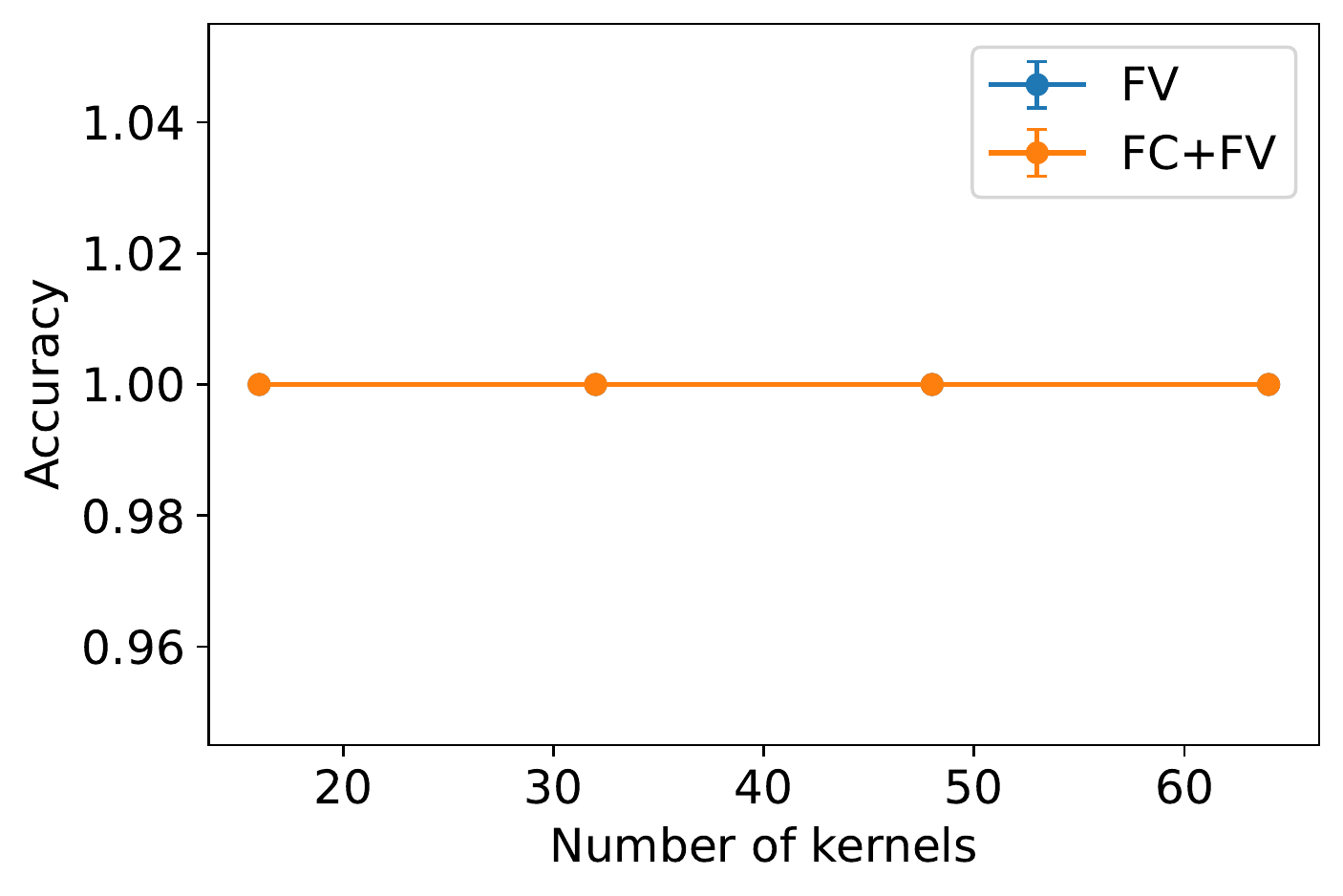}\\
	\end{tabular}
	\begin{tabular}{c}
		UIUC\\
		\includegraphics[width=.45\textwidth]{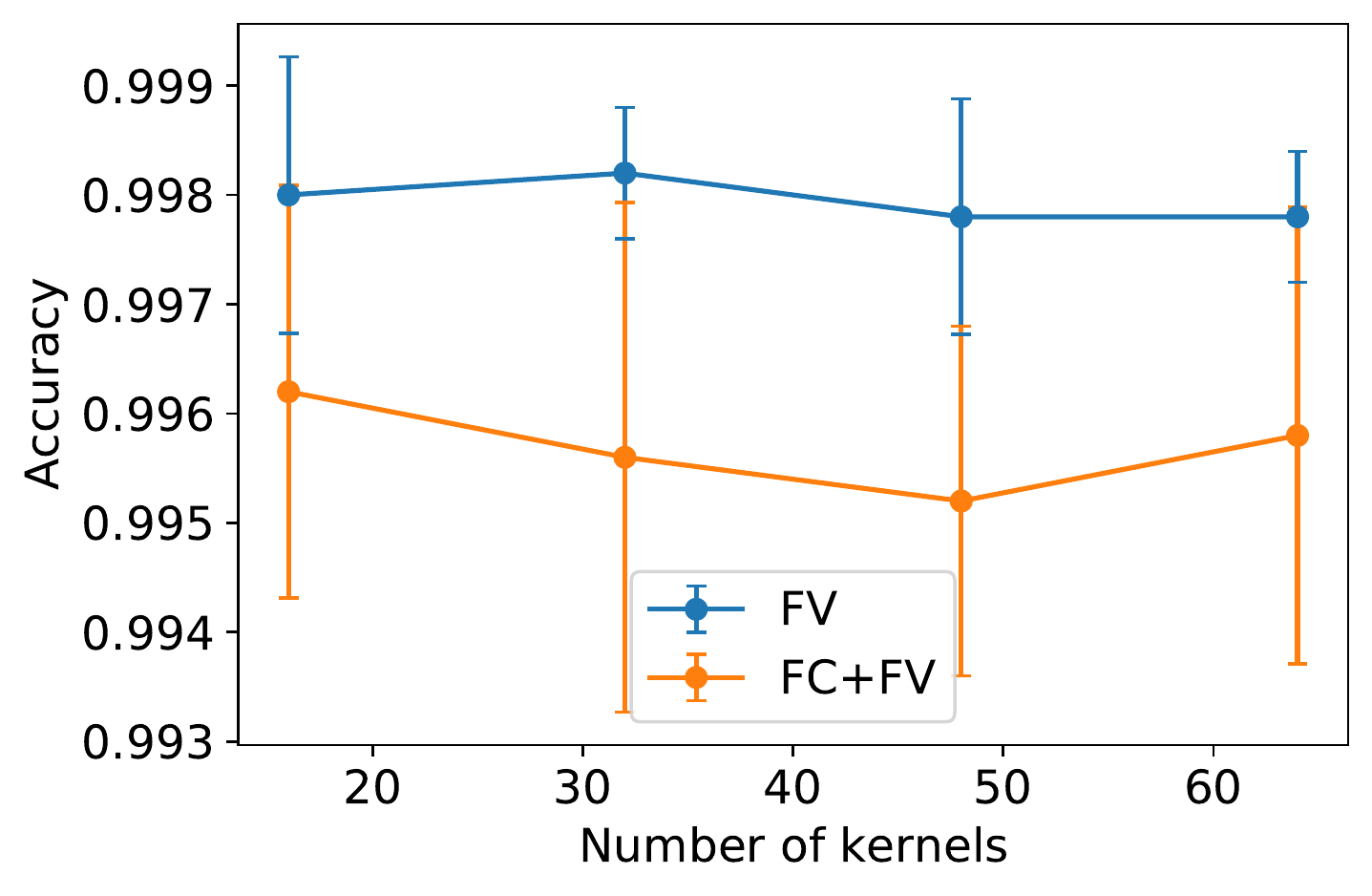}
	\end{tabular}
    \caption{Variation of accuracy of our method according to the number of Gaussian distributions (kernels) used in GMM. Line colors describe which information was used for training the classifier. Error bars indicate the standard deviation of classification accuracy.}
    \label{fig:kernels}
\end{figure}

In this section we present the results obtained from the experiments described in Section~\ref{sec:experiments}. We show how they accomplished to verify the effectiveness of the proposed methodology in texture classification. As mentioned in Section~\ref{sec:experiments}, the accuracy of our method can be affected by the number of Gaussian distributions that we choose to model $u_{\lambda}$. Those distributions are also called number of \emph{kernels} or \emph{visual words}. In our tests, we call this hyperparameter \emph{number of kernels}. We used $16$, $32$, $48$ and $64$ kernels in benchmark tests. The results are show in Figure~\ref{fig:kernels}. We observed very little variation of accuracy in the case of FMD and UMD, but a significant increase in accuracy as we increase the number of kernels for KTH-TIPS and DTD. Thus, using 64 kernels seems to be a good choice for all the databases tested. An improvement with increasing kernels is expected, as the greater the number of Gaussian distributions, the better it can model the underlying distribution that generates the local features. However, the number of Gaussian distributions should not be very large given the limited availability of data to train the GMM algorithm and computational costs.

\begin{figure}[!htpb]
	\centering
	\begin{tabular}{cc}
		KTH-TIPS & FMD\\
		\includegraphics[width=.45\textwidth]{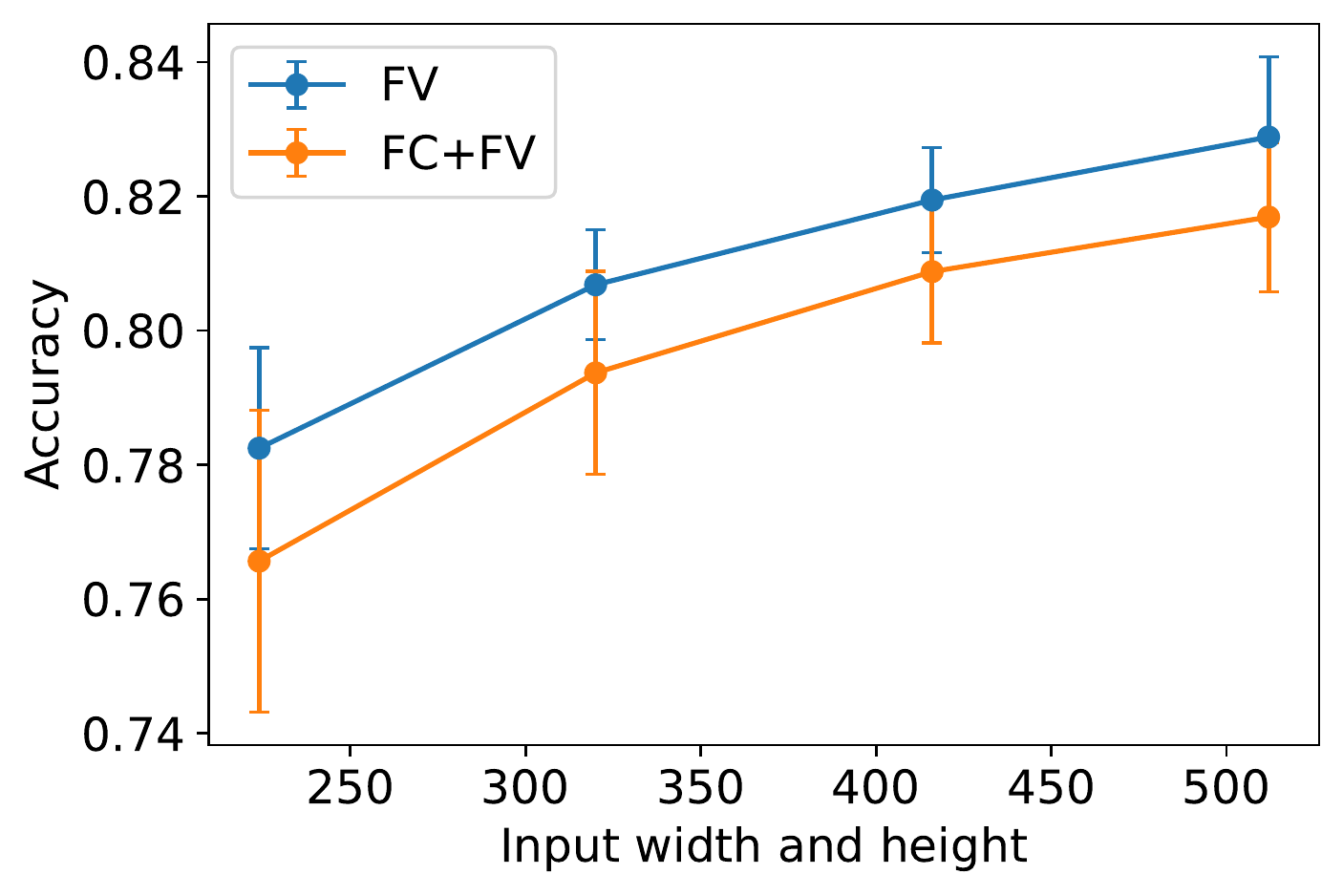} &
		\includegraphics[width=.45\textwidth]{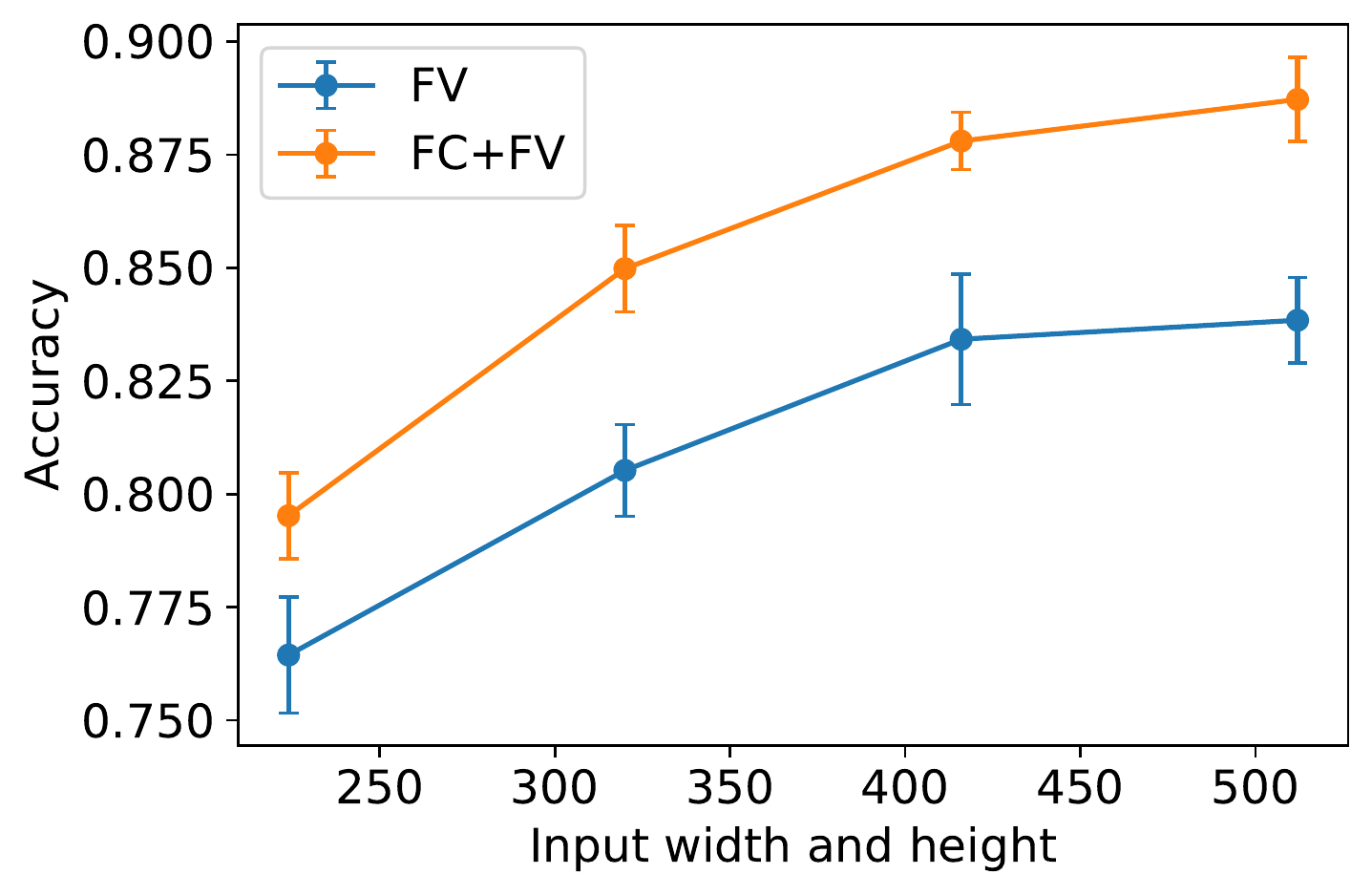}\\
	\end{tabular}
	\begin{tabular}{cc}
		DTD & UMD\\
		\includegraphics[width=.45\textwidth]{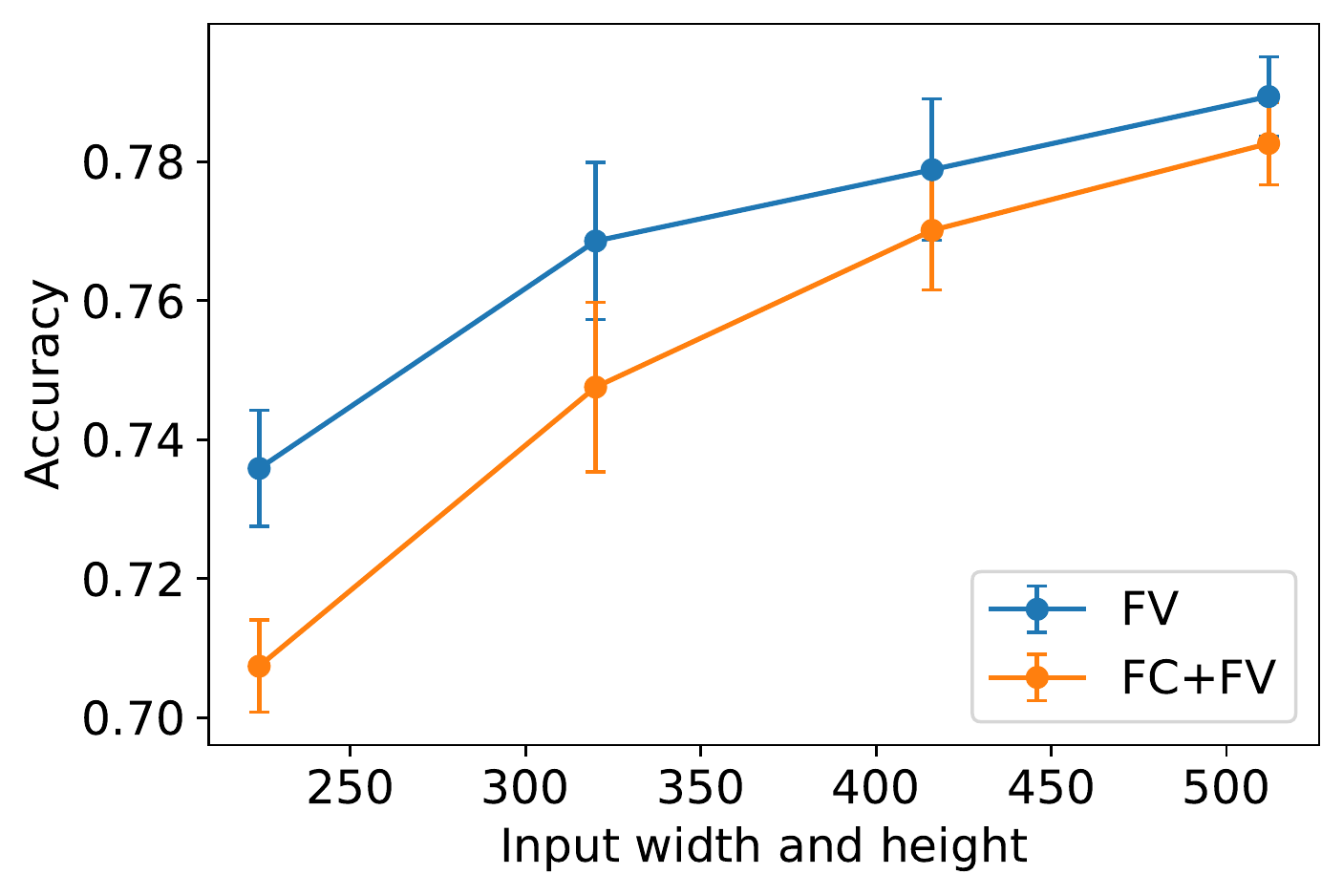} &
		\includegraphics[width=.45\textwidth]{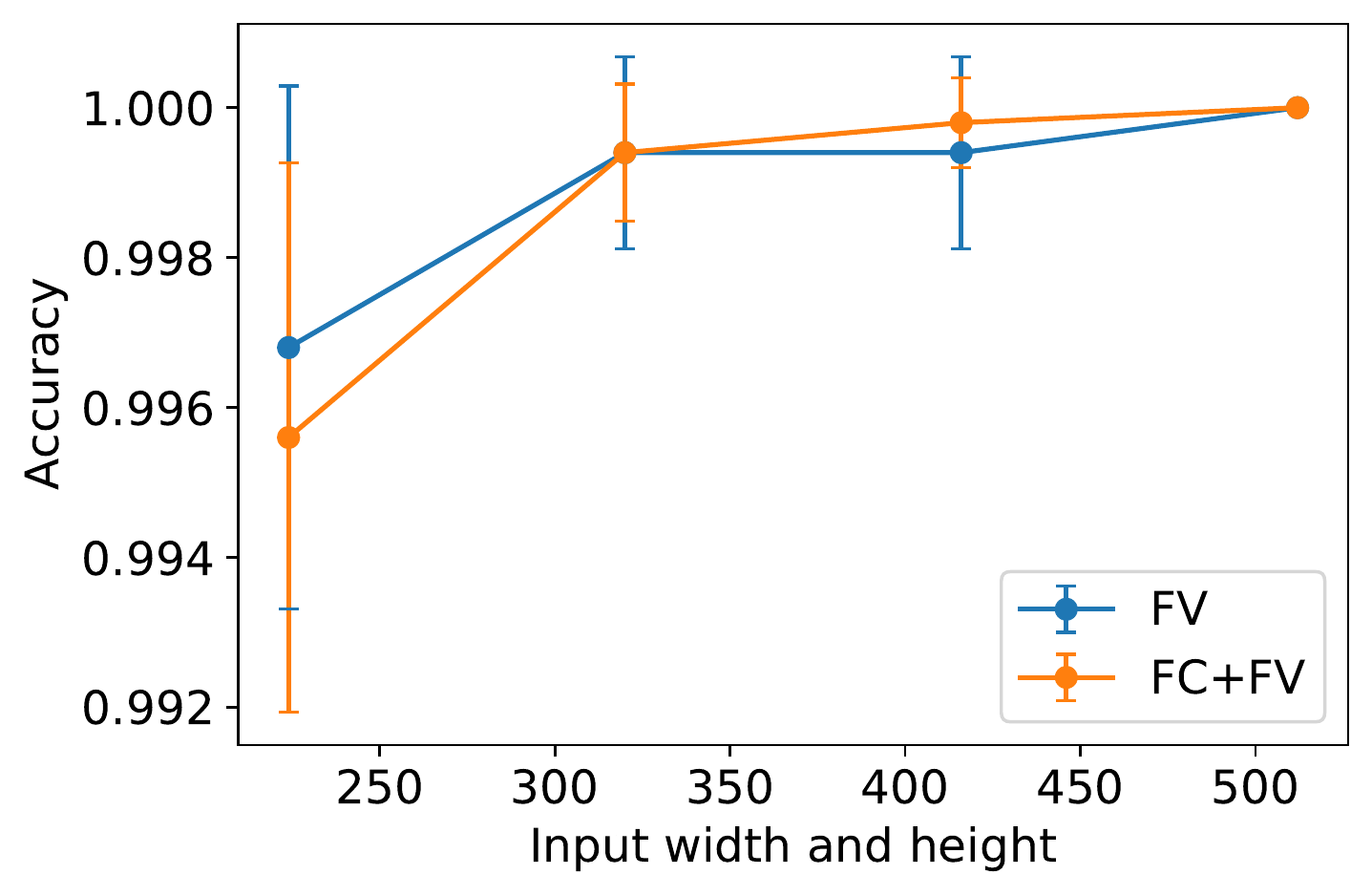}\\
	\end{tabular}
	\begin{tabular}{c}
		UIUC\\
		\includegraphics[width=.45\textwidth]{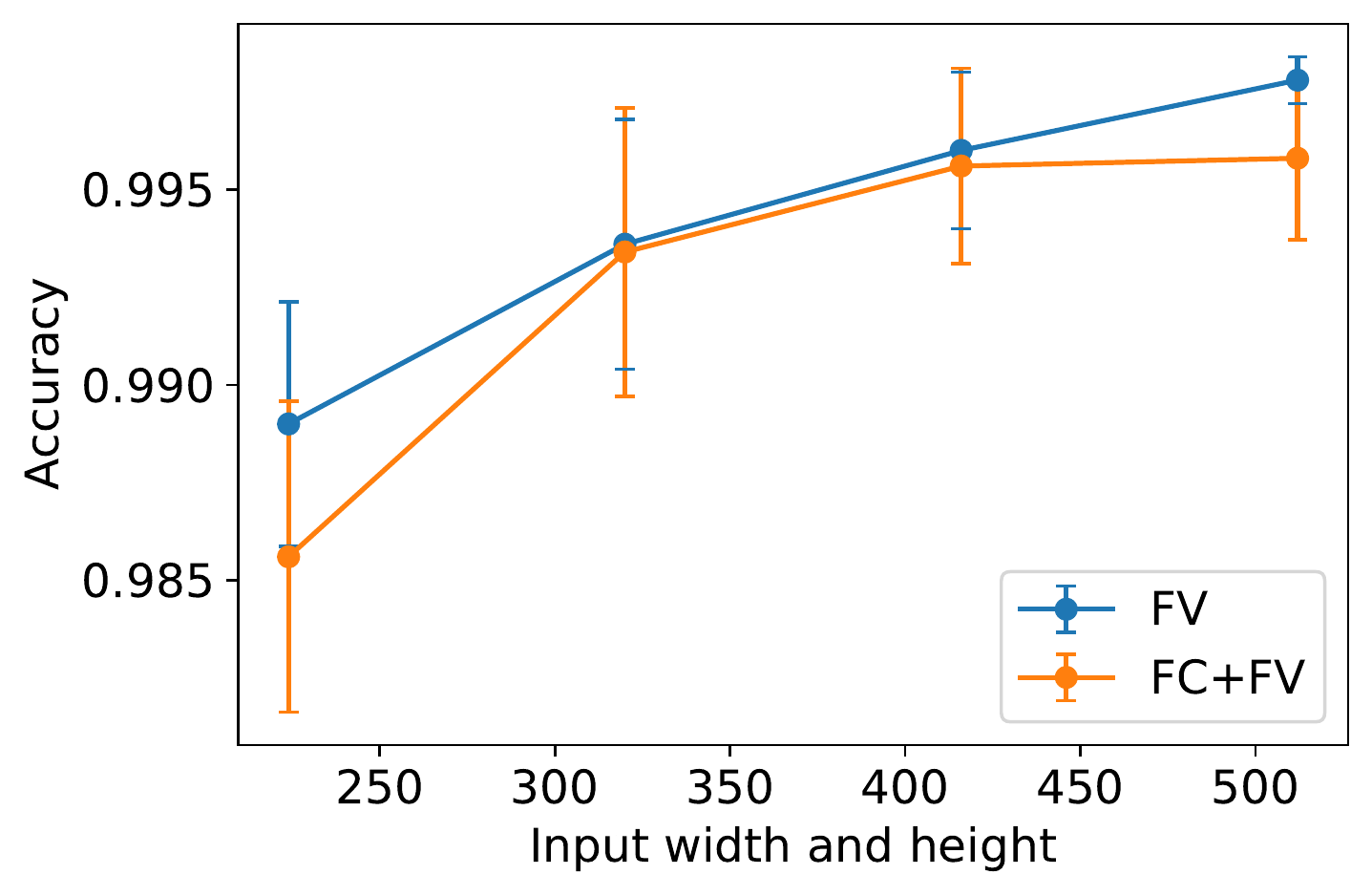}
	\end{tabular}
    \caption{Accuracy of our method according to the resolution of the image used as input to the CNN. Line colors describe the feature vectors used for training the classifier. Error bars indicate the standard deviation of classification accuracy.}
	\label{fig:resolution}
\end{figure}

The second hyperparameter of our model is the resolution of the input image. This resolution is directly proportional to the number of local features, which is linked to the performance of the GMM algorithm. We evaluated our model on the benchmark databases starting with $224 \times 224$, the size used by the CNN architecture for training. We increase the resolution linearly up to $512 \times 512$ to show its effect on GMM algorithm. The results of this variation are shown in Figure~\ref{fig:resolution}. As expected, the increase in image resolution improved accuracy across all databases. This improvement is not only due to the number of local features, but also to how specific a local feature is. If the resolution is too small, information from small regions in an image may be lost. A condition for the use of generative models to be beneficial for accuracy is that local features must describe small regions rather than large ones.

In our third experiment, we show how our method behaves in different network architectures. We chose architectures that result in local features with similar dimensions. We tested our method in 
\begin{itemize}
    \item EfficientNet-B5, where local feature dimension $D=176$;
    \item EfficientNetV2-s \cite{tan2021efficientnetv2}, where $D=160$;
    \item ResNet34 \cite{he2016deep}, where $D=256$.
\end{itemize}
As shown in Table~\ref{tab:arch}, the best accuracy for KTH-TIPS2-b was achieved in EfficientNet-B5 while EfficientNetV2-s achieved better accuracy in FMD and DTD. Very deep ResNet, VGG \cite{simonyan2014very}, DenseNet \cite{huang2017densely} would increase greatly local feature dimensions and consequently the computational cost of GMM algorithm. For example, in DenseNet-161 local feature dimension is $D=2048$.

\begin{table}
    \centering
    \caption{Behavior of our proposed method in different CNN architectures. Column ``Method'' describes the feature vectors that were used to train the classifier.}
    \begin{tabular}{l l c c c}
        \hline
        Dataset & Method & EfficientNet-B5 & EfficientNetV2-s & ResNet34 \\
        \hline
        KTH-TIPS & FV & $82.9 \pm 1.2$ & $80.6 \pm 1.3$ & $79.2 \pm 1.7$ \\
         & FV+FC & $81.7 \pm 1.1$ & $79.5 \pm 1.5$ & $77.3 \pm 3.7$ \\
        FMD & FV & $83.8\pm 0.9$ & $85.7 \pm 1.2$ & $82.2 \pm 1.4$ \\
         & FV+FC & $88.7\pm 0.9$ & $88.9 \pm 1.0$ & $83.5 \pm 1.2$ \\
         DTD & FV & $78.9\pm 0.6$ & $79.3 \pm 0.6$ & $76.2 \pm 0.3$ \\
         & FV+FC & $78.3\pm 0.6$ & $77.8 \pm 1.1$ & $71.3 \pm 0.7$ \\
         UMD & FV & $100 \pm 0.0$ & $100 \pm 0.0$ & $99.9 \pm 0.1$ \\
         & FV+FC & $100 \pm 0.0$ & $100 \pm 0.0$ & $99.9 \pm 0.1$\\
         UIUC & FV & $99.8 \pm 0.1$ & $99.8 \pm 0.1$ & $99.8 \pm 0.1$ \\
         & FV+FC & $99.6 \pm 0.2$ & $99.7 \pm 0.2$ & $99.7 \pm 0.3$ \\
         \hline
    \end{tabular}
    \label{tab:arch}
\end{table}

\begin{figure}[!htpb]
	\centering
	\begin{tabular}{cc}
		FC without normalization & FC with normalization\\
		\includegraphics[width=.45\textwidth]{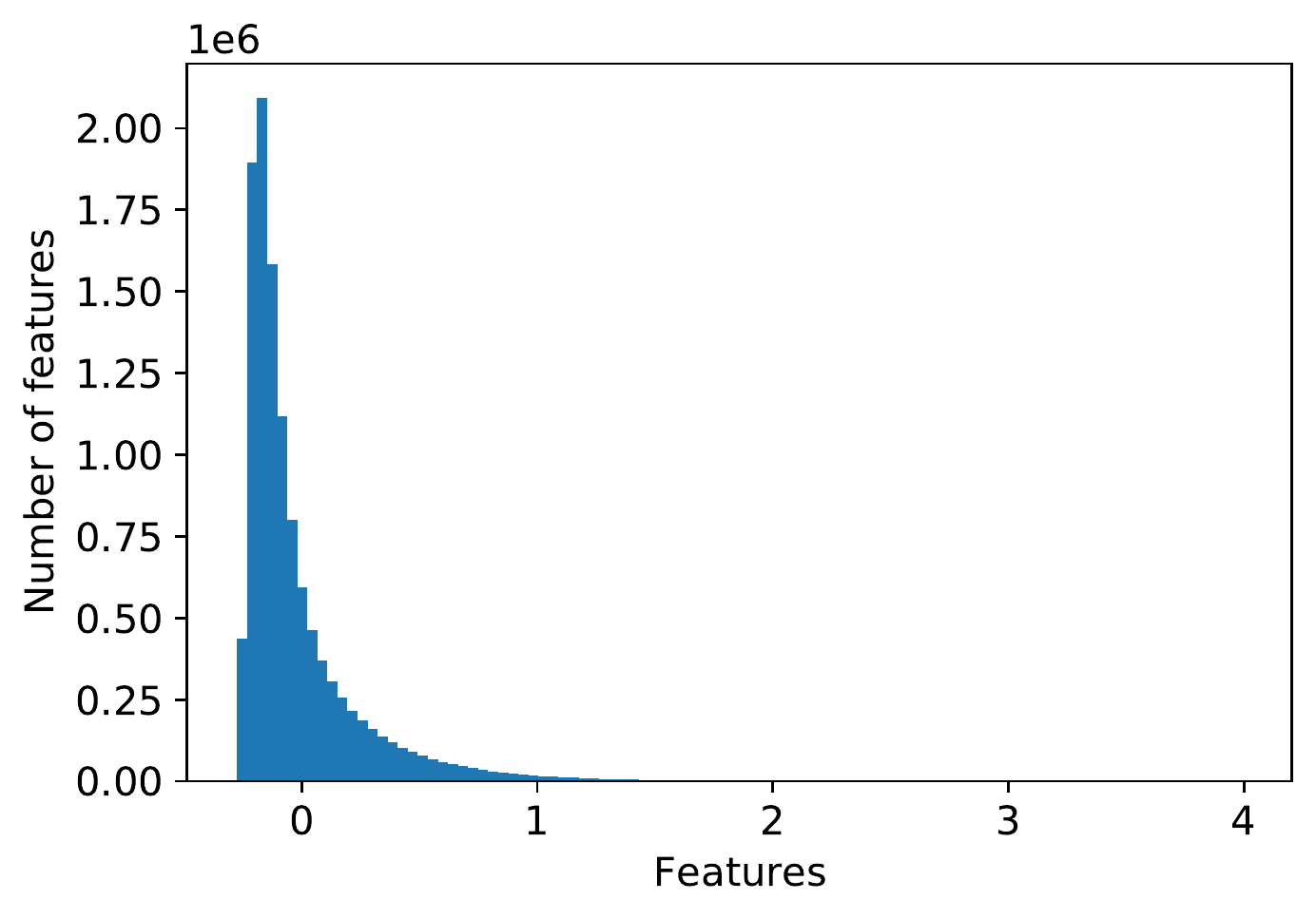} &
		\includegraphics[width=.45\textwidth]{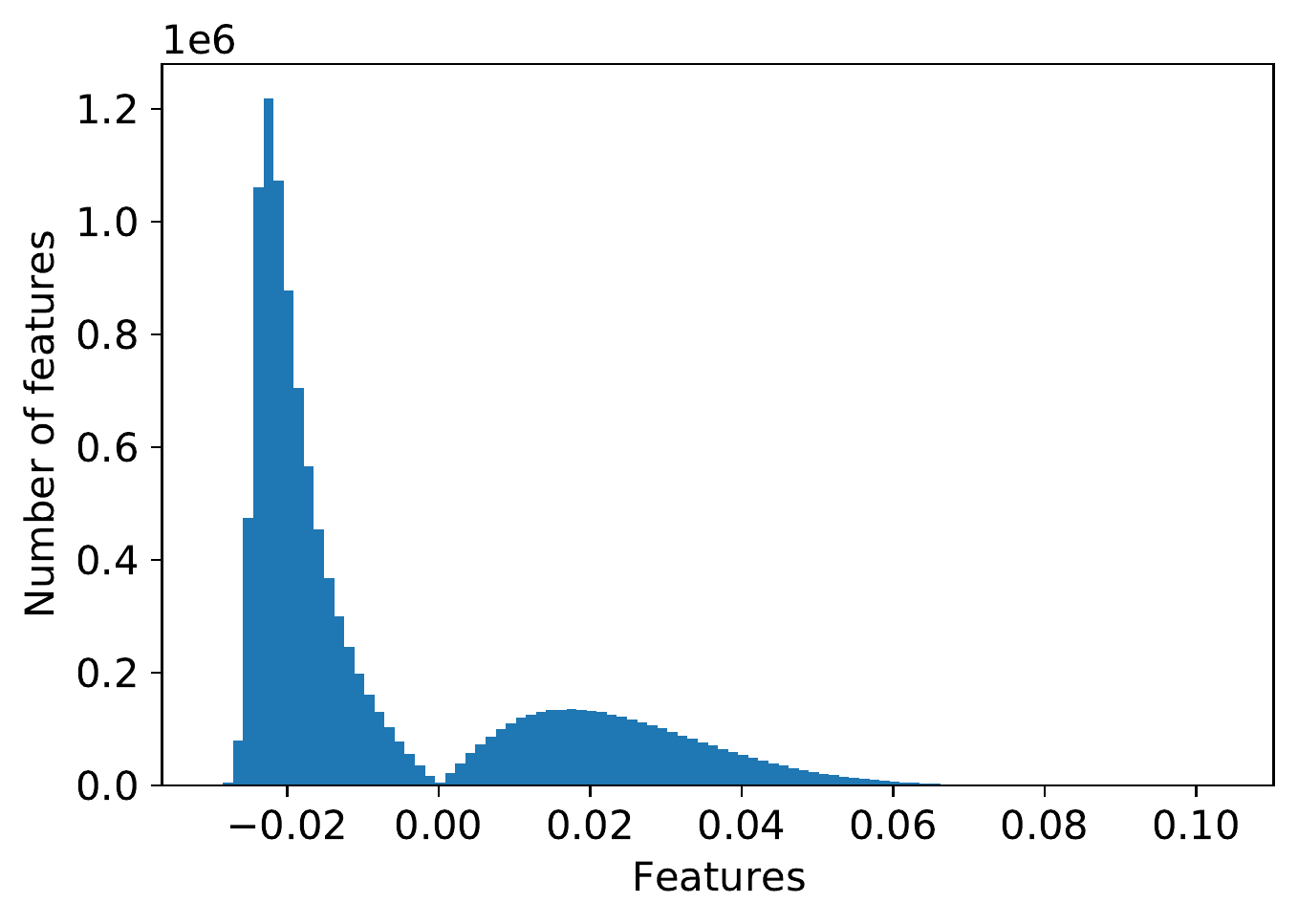}\\
	\end{tabular}
    \caption{Histograms of FC calculated for DTD database before and after applying normalization given by Equations~(\ref{eq:norm1}) and (\ref{eq:norm2}).}
    \label{fig:hist}
\end{figure}

Moreover, we verify the impact of the power and $L_{2}$ normalization applied to FC. All FC features used for evaluation are extracted from the architecture EfficientNet-B5. In Figure~\ref{fig:hist} we show the impact of the normalization of the distribution of FC elements for DTD database. For the other databases, the effect is similar. The normalization affects the format of the distribution and increases data sparsity. In Table~\ref{tab:fc} we show the effect of normalization on accuracy considering exclusively FC classification. In general, it helped increasing accuracy mean or reducing standard deviation. Most notorious result can be seen in DTD database, where both effects are present, while normalization had no impact in UMD.

\begin{table}[ht]
    \centering
    \caption{Normalization impact on accuracy for benchmark databases. In this experiment, SVM was used to classify FC, in the first case with no transformation applied, in the second case, applying the normalization proposed in Section~\ref{sec:classification}.}
    \begin{tabular}{l c c}
        \hline
        Database & Without Normalization & With Normalization\\
        \hline
        KTH-TIPS & $78.8 \pm 1.9$ & $79.1 \pm 1.9$ \\
        FMD & $86.6 \pm 1.2$ & $86.8 \pm 0.9$ \\
        DTD & $72.9 \pm 0.8$ & $73.3 \pm 0.7$ \\
        UMD & $100 \pm 0.0$ & $100 \pm 0.0$ \\
        UIUC & $99.3 \pm 0.3$ & $99.2 \pm 0.2$ \\
        \hline
    \end{tabular}
    \label{tab:fc}
\end{table}

\begin{figure}[!htpb]
	\centering
	\begin{tabular}{cc}
		KTH-TIPS & FMD\\
		\includegraphics[width=.45\textwidth]{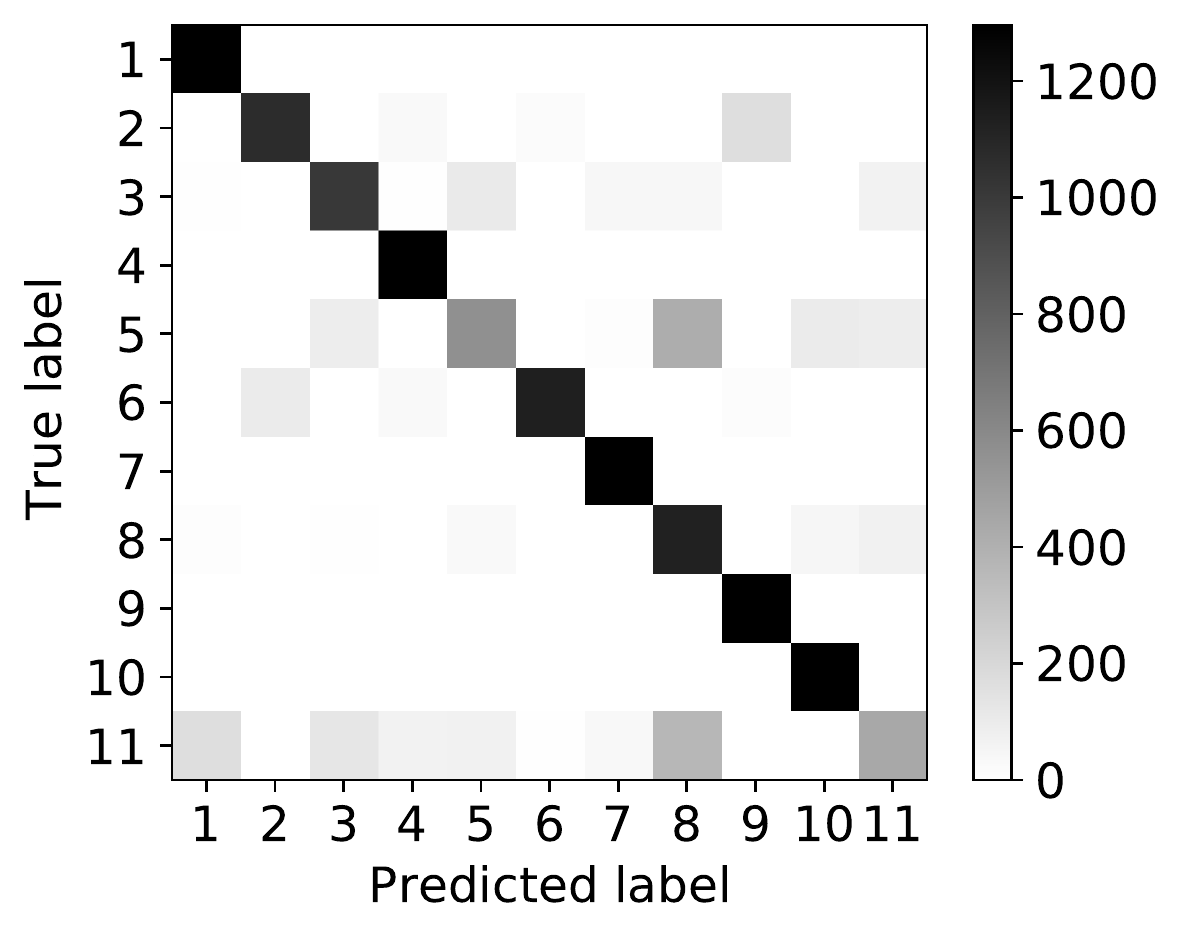} &
		\includegraphics[width=.45\textwidth]{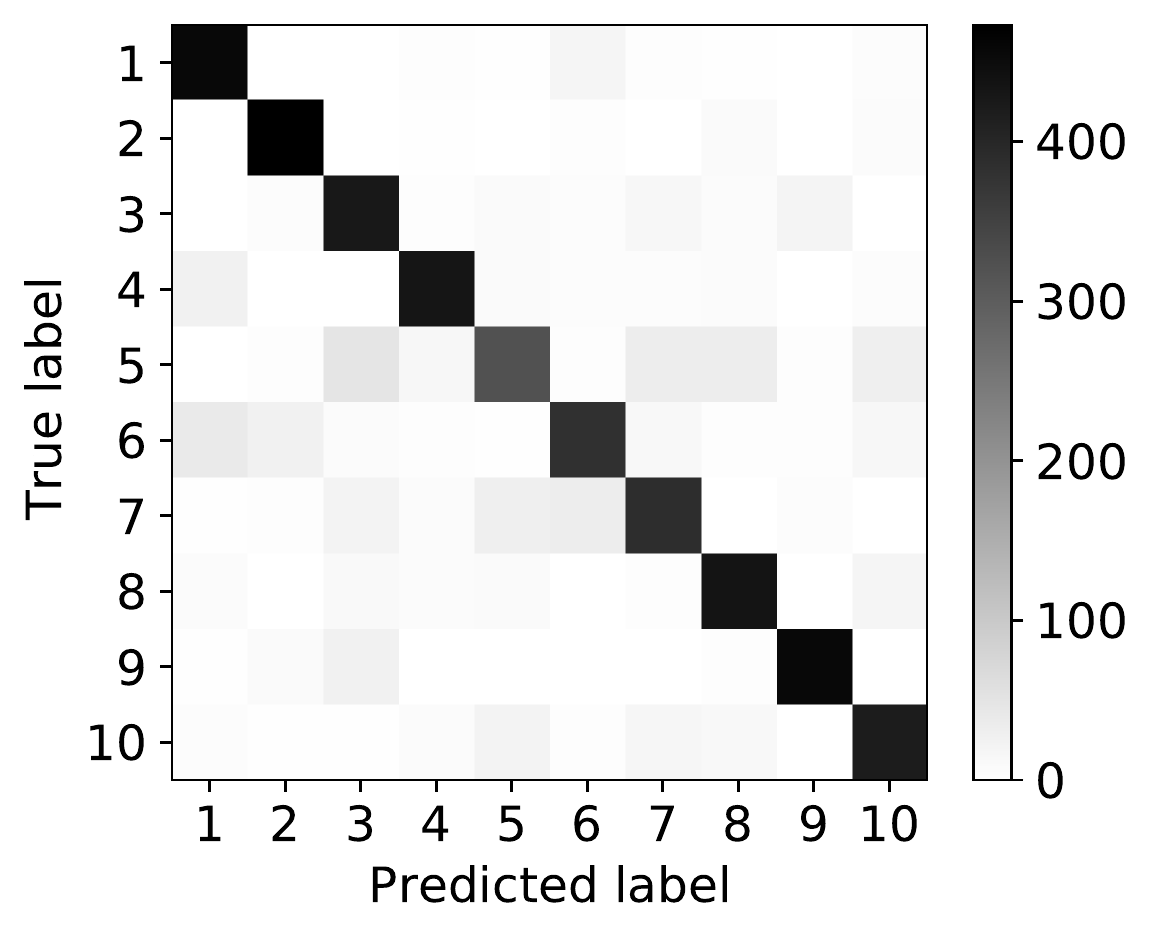}\\
	\end{tabular}
	\begin{tabular}{cc}
		DTD & UMD\\
		\includegraphics[width=.45\textwidth]{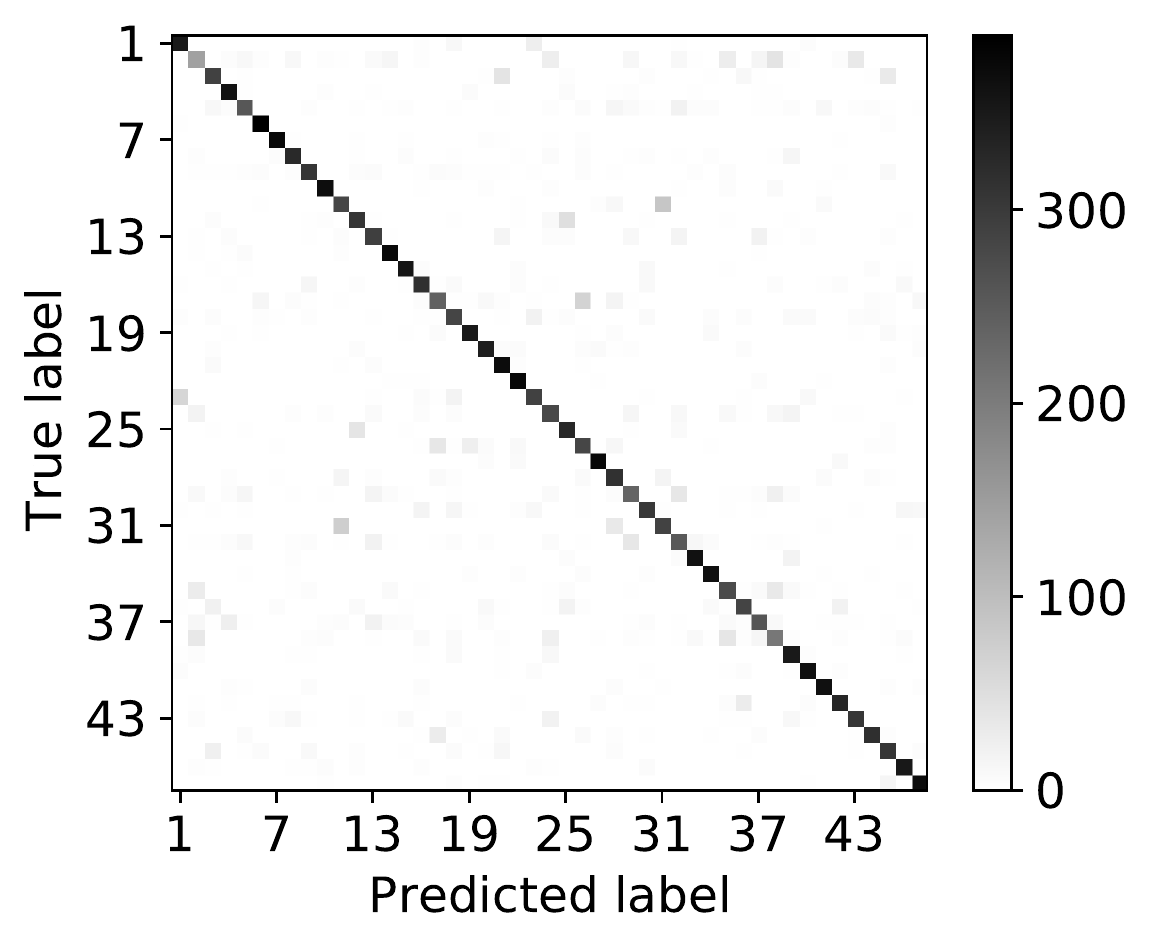} &
		\includegraphics[width=.45\textwidth]{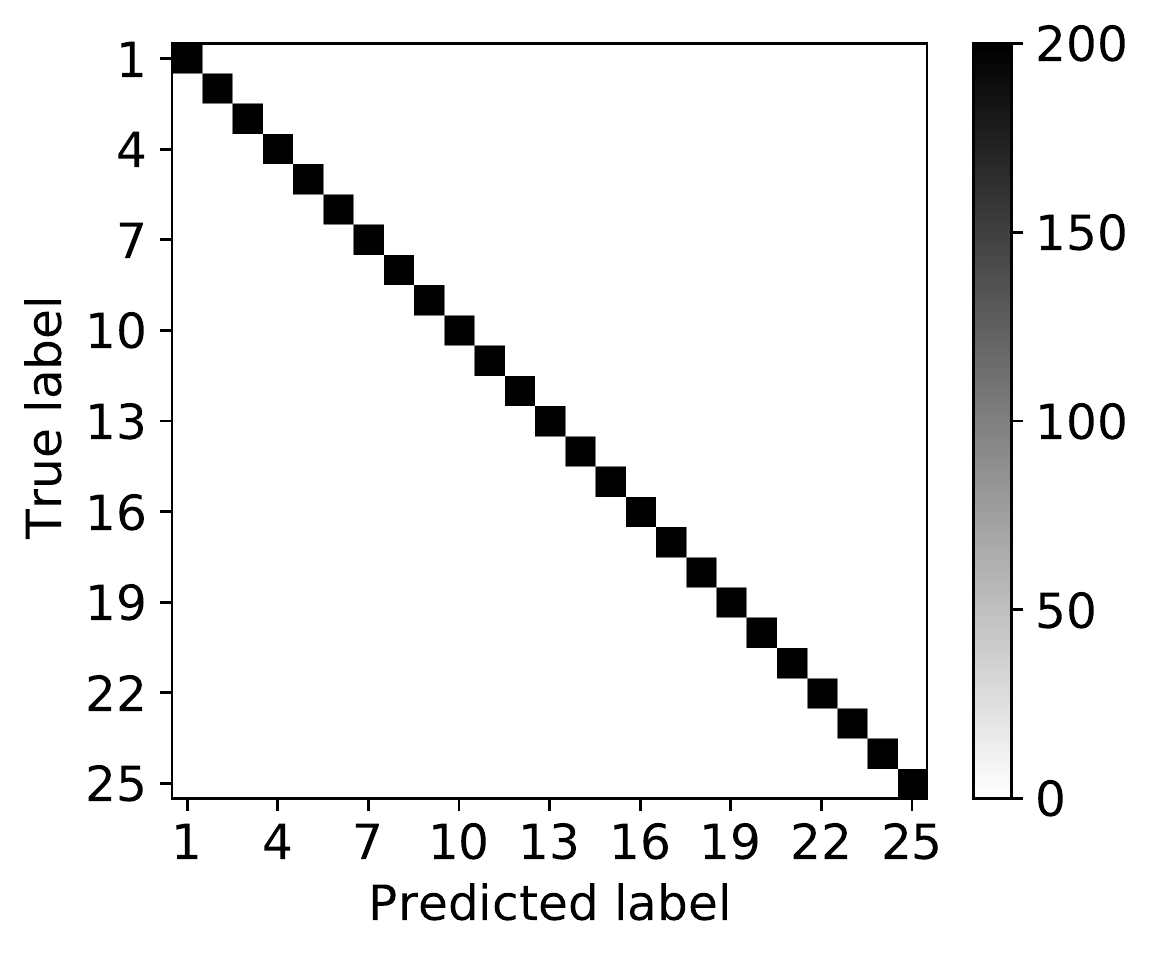}\\
	\end{tabular}
	\begin{tabular}{c}
		UIUC\\
		\includegraphics[width=.45\textwidth]{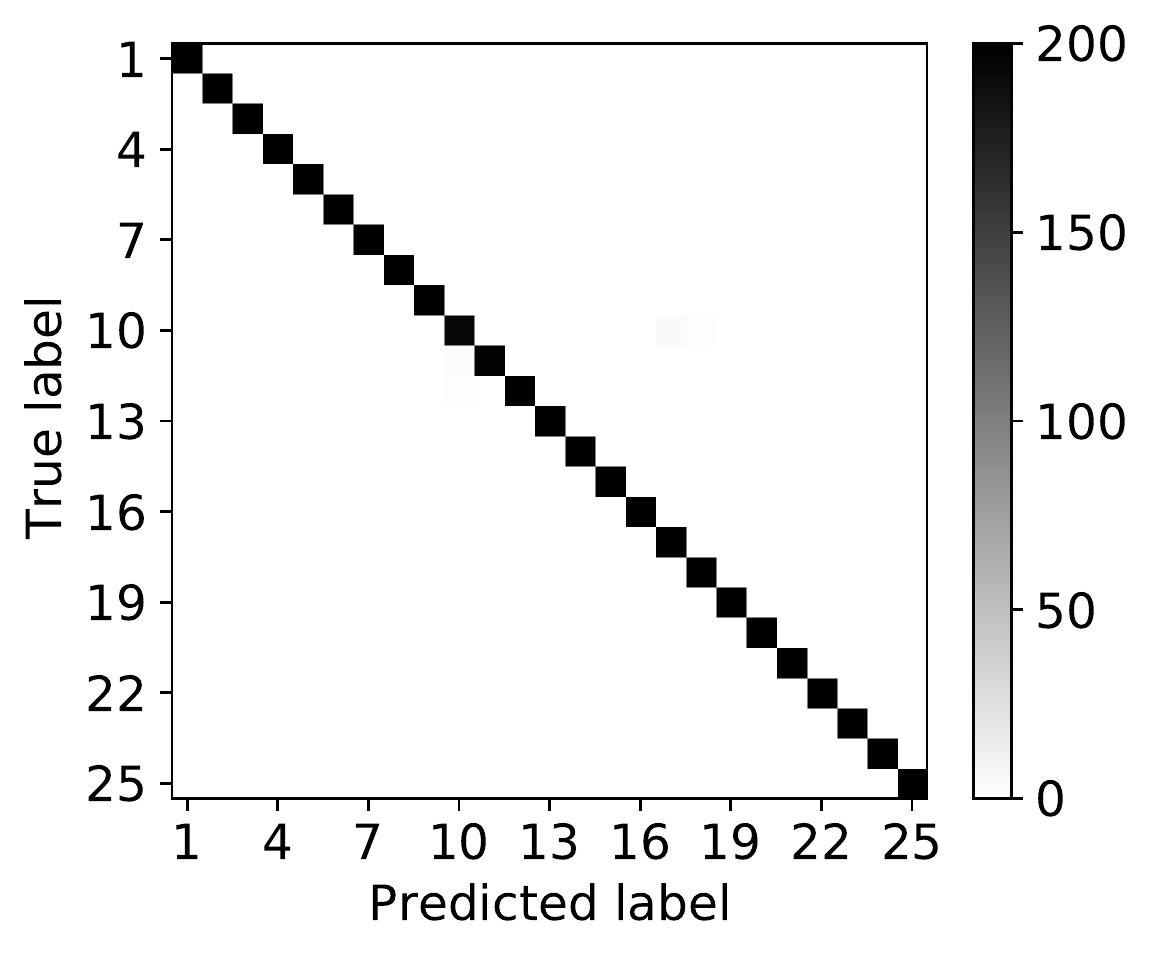}
	\end{tabular}
	\caption{Confusion matrices computed with our classifier trained with FV, which encodes information from multiple convolutional layers. These matrices show the number of images that were assigned to certain class. No confusion means all images are labeled their true class, i.e., all diagonal elements are black and the remaining elements are white.}
	\label{fig:matrices}
\end{figure}

For all the following results, the architecture used is the EfficientNet-B5, the input image resolution is $512 \times 512$ and the number of kernels is $64$. In Figure~\ref{fig:matrices} we detail how our method behaves in the benchmark databases by showing how much confusion is presented in each database. 

In KTH-TIPS, most noticeable problems are the classification of examples from class $5$ (cotton) and class $11$ (wool). In the case of cotton, its mostly confused with class $8$ (linen), although there are certain confusion also with classes $3$ (corduroy), $10$ (wood) and $11$. In the case of wool, its mostly confused with linen, but there is also confusion with classes $1$ (aluminium foil) and $3$. Interestingly, most part of the confusion is among textile textures, which are indeed challenging to classify, given that they can have a similar pattern.

In FMD, our model had most problems distinguishing class $5$ (metal) from other classes, confusing it with classes $3$ (glass), $7$ (plastic), $8$ (stone) and $10$ (wood). The presence of confusion in this case could be explained by the fact that objects made out from these materials can present a similar shape or color to metallic objects.

In DTD, the most notorious classification problem of our model is perceived in class $2$ (blotchy), where less than $50\%$ of samples are correctly classified. These samples are mostly mistaken by classes $38$ (stained) and $43$ (veined). The confusion between blotchy and stained was expected, as images from both classes are very similar. Also, the edges between botched and non-blotched regions in an image can be mistakenly interpreted as veins, what could explain confusion with class $43$. 
No significant confusion can be observed in UIUC and UMD databases.

In Table~\ref{tab:result}, we list the accuracy of several methods in the literature of texture recognition compared with the proposed approach. Our proposed method using only FV outperforms other modern deep learning approaches in KTH-TIPS, DTD and UMD. The accuracy we achieved in DTD is, as far as we know, the best result available in the literature. Furthermore, our proposed method combining FC and FV is able to, to the best of our knowledge, achieve state-of-art performance in FMD database.

\begin{table}[ht]
    \centering
    \caption{Accuracy comparison with other methods in literature. Our proposed method is named here Multilayer-FV and Multilayer-FV+FC. All results shown are obtained directly from the original paper of each method. Non-published results are represented by dashes.}
    \begin{tabular}{l c c c c c}
        \hline
        Method & KTH-TIPS & FMD & DTD & UMD & UIUC \\
        \hline
        FV-VGGVD \cite{cimpoi2016deep} & 81.8 & 79.8 & 72.3 & 99.9 & 99.9 \\
        SIFT-FV \cite{cimpoi2016deep} & 81.5 & 82.2 & 75.5 & 99.9 & 99.9 \\
        LFV \cite{song2017locally} & 82.6 & 82.1 & 73.8 & - & - \\
        VisGraphNet \cite{florindo2021visgraphnet} & 75.7 & 77.3 & - & 98.1 & 97.6 \\
        Non-Add Entropy \cite{florindo2021using} & - & 77.7 & - & 98.8 & 98.5 \\
        Xception + SIFT-FV \cite{jbene2019fusion} & - & 86.1 & 75.4 & - \\
        Residual Pooling \cite{mao2021deep} & - & 85.7 & 76.6 & - & - \\
        FENet \cite{xu2021encoding} & - & 86.7 & 74.2 & - & - \\
        CLASSNet \cite{chen2021deep} & - & 86.2 & 74.0 & - & - \\
        \hline
        Multilayer-FV & $\mathbf{82.9}$ & 83.8 & $\mathbf{78.9}$ & 100.0 & 99.8 \\
        Multilayer-FV+FC & 81.7 & $\mathbf{88.7}$ & 78.3 & 100.0 & 99.6 \\
        \hline
    \end{tabular}
    \label{tab:result}
\end{table}

\begin{figure}[!htpb]
	\centering
	\begin{tabular}{cc}
		Kernel variation & Image resolution variation\\
		\includegraphics[width=.45\textwidth]{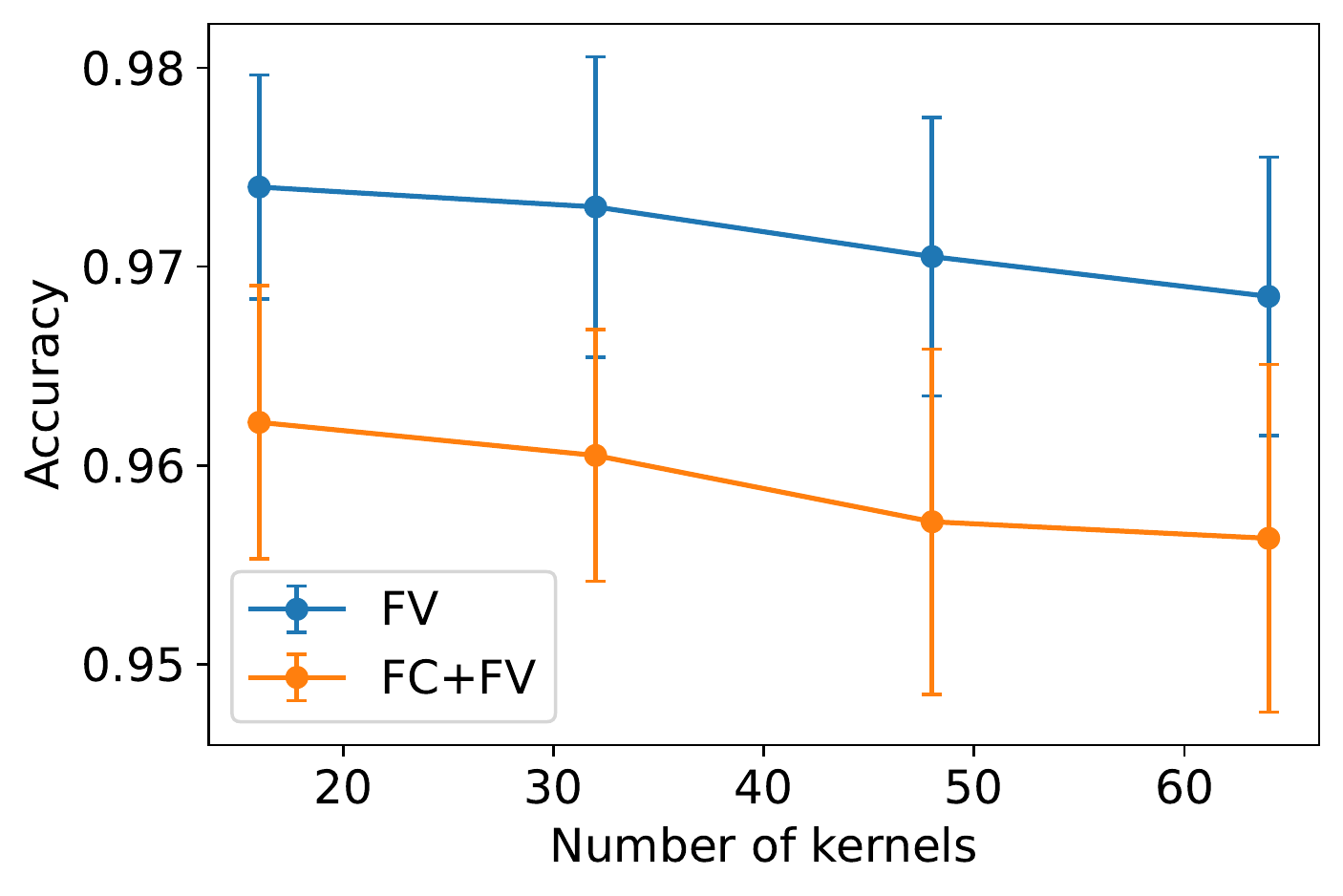} &
		\includegraphics[width=.45\textwidth]{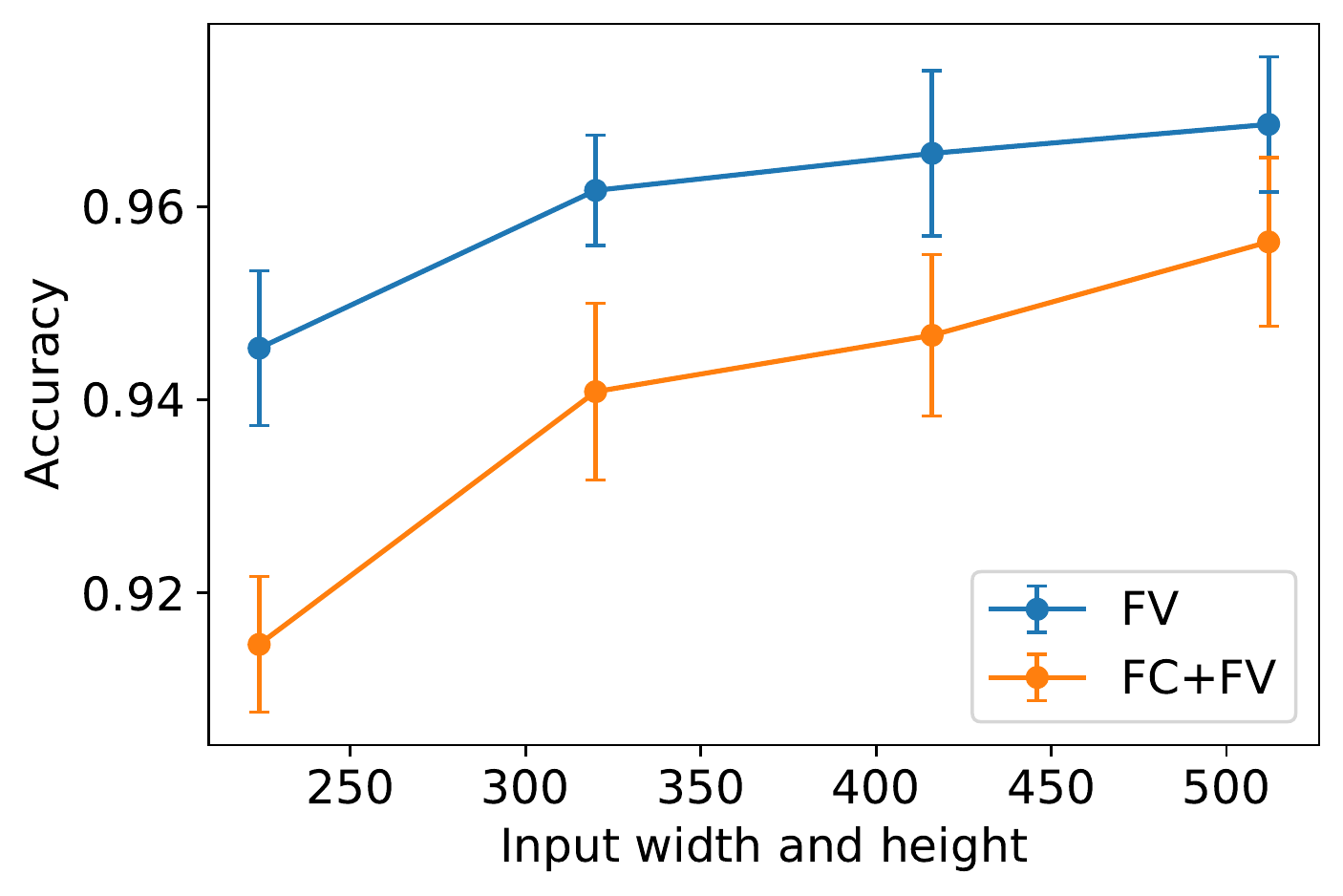}\\
	\end{tabular}
	\caption{Accuracy of our proposed method for different number of kernels (Gaussian distributions in GMM) and different resolutions of the image provided to the CNN.}
	\label{fig:1200hyper}
\end{figure}

Finally, we apply our model to the classification task of Brazilian plant species. 
We first evaluate the impact of hyperparameter change on the database. In Figure~\ref{fig:1200hyper}, we show that the increase in number of kernels affects negatively the accuracy of classification. 
This is probably caused by the low availability of data in order to train the GMM algorithm for a greater number of Gaussian distributions. 
We can also see that increasing the resolution of the input image affects accuracy positively as in all other databases tested.

\begin{figure}
\centering
\includegraphics[width=.5\linewidth]{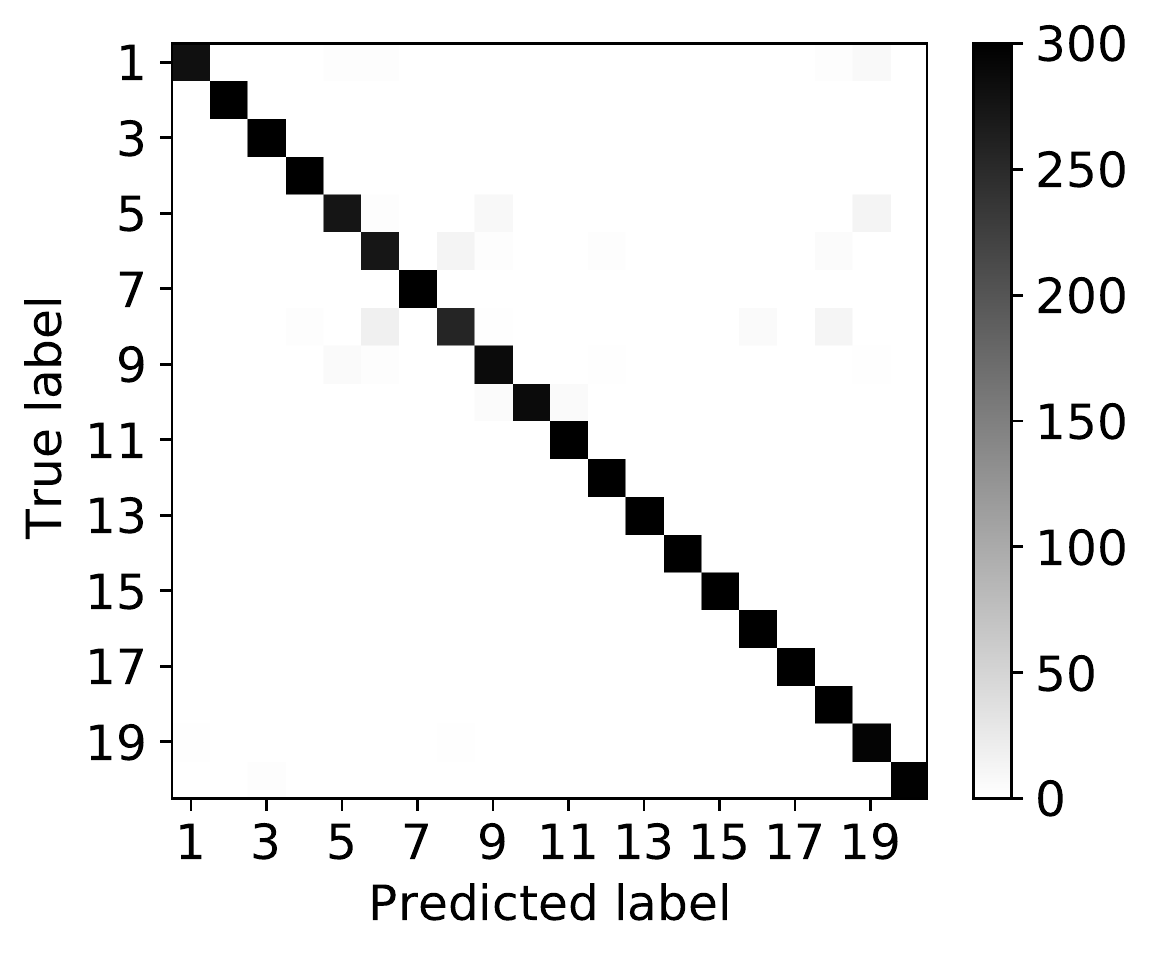}  
\caption{Confusion matrix for our method in 1200Tex. This matrix was computed using a classifier trained exclusively with information from FV descriptors.}
\label{fig:matrix1200}
\end{figure}

For the particular task of evaluating and comparing the behavior of our model in 1200Tex database, we use 16 Gaussian distributions to model $u_{\lambda}$.
This is done because, as shown in Figure~\ref{fig:1200hyper}, our method behaves better with few distributions in this case. 
In Figure~\ref{fig:matrix1200} we note that there is not much confusion when classifying the plant species. The classes that our model has most problems classifying are $8$, wrong labelling around $14.5\%$ of samples, and $5$ and $6$, in both around $8.5\%$ of samples are confused with other classes.
Class $8$, which presents a green leaf mostly dotted with few veins, is confused with classes $6$, which is also veined, and $18$, which is dotted and veined. Confusion is this case can be generated when examples from $8$ have more veined areas than dotted, being wrongly labelled according to the proportions between those areas.
Class $6$ is mostly confused with class $8$, which could be due to image or leaf imperfection in some examples from class $6$ being interpreted as dotted regions.
Class $5$, which is mostly smooth with few grained areas, is confused with class $19$, which is mostly grained. This confusion can be generated when focus is given to grained areas in images from class $5$.


In Table~\ref{tab:1200tex} we list the accuracy of the best previous results on 1200Tex database that we found in literature, in comparison with our proposal. Here, the usage of our methodology made a huge difference in accuracy, scoring a result $9\%$ better than the second best method (Non-Add Entropy). In fact, up to our knowledge, this result sets a new state-of-the-art accuracy on this database.

\begin{table}
    \centering
    \caption{Comparison of accuracy in 1200Tex database with other methods in literature. Our method is presented as Multilayer-FV. All results were obtained directly from the literature. When results were not found in the original paper, additional reference was given to where the result was taken from.}
    \begin{tabular}{l c}
        \hline
         Method & Accuracy (\%) \\
         \hline
         SIFT+BOVW \cite{cimpoi2016deep} & 86.0 \cite{silva2021fractal} \\
         FV-VGGVD \cite{cimpoi2016deep} & 87.1 \cite{florindo2021using} \\
         Fractal \cite{silva2021fractal} & 86.3 \\
         VisGraphNet \cite{florindo2021visgraphnet} & 87.4 \\
         Non-Add Entropy \cite{florindo2021using} & 88.5 \\
         \hline
         Multilayer-FV & 97.4 \\
         \hline
    \end{tabular}
    \label{tab:1200tex}
\end{table}

\section{Conclusions}
\label{sec:conclusions}

In this work, we proposed and investigated the use of local features extracted from multiple convolutional layers and how this improves classification using Fisher Vector. More precisely, we computed the Fisher Vector on local features extracted from the last two convolutional layers and used it as texture descriptor.

We evaluated the performance of our method in visual texture classification, both in benchmark databases and in a practical problem of identifying plant species. In both situations, our method presented a significant improvement over other methods in the literature and reached competitive accuracy with the state-of-the-art. This good performance can be explained by some points. One of them is the use of a mixture of more generalist features extracted from earlier convolutional layers and more domain-specific features extracted from later layers. The second factor is the adjustment of the input image to a resolution higher than the CNN standard, which affects both the number of local features and the specificity of a local feature to a given area of the input image. A last point is the use of normalization in the FC descriptor, which resulted in improvement of the method by combining the FV and FC descriptors.

The results expressed here also suggest that a combination of outputs from previous layers might be beneficial for classification accuracy. Also, different ways of combining local features from multiple layers may help to preserve better information from later layers and is a topic for future investigation.

\section*{Acknowledgements}
L. O. L. gratefully acknowledges the financial support of Coordination for the Improvement of Higher Education Personnel, Brazil (CAPES) (Grant \#1796018).
J. B. F. gratefully acknowledges the financial support of S\~ao Paulo Research Foundation (FAPESP) (Grant \#2020/01984-8) and from 
National Council for Scientific and Technological Development, Brazil (CNPq) (Grants \#306030/2019-5 and \#423292/2018-8).


\end{document}